\documentclass[sigconf,nonacm]{acmart}
\usepackage[utf8]{inputenc}
\usepackage[font=small,labelfont=bf]{caption}
\usepackage{amsmath}
\usepackage{gensymb}
\usepackage{subfigure}
\usepackage{svg}
\usepackage{mathtools}
\usepackage{siunitx}

\AtBeginDocument{%
  \providecommand\BibTeX{{%
    \normalfont B\kern-0.5em{\scshape i\kern-0.25em b}\kern-0.8em\TeX}}}

\begin{document}

\title{Bridging the Gap: Gaze Events as Interpretable Concepts to~Explain~Deep~Neural~Sequence~Models}

\renewcommand{\shorttitle}{Bridging the Gap: Gaze Events as Interpretable Concepts to Explain Deep Neural Sequence Models}


\author{Daniel G. Krakowczyk}
\email{daniel.krakowczyk@uni-potsdam.de}
\orcid{0009-0009-5100-0733}
\affiliation{%
  \institution{University of Potsdam}
  \city{Potsdam}
  \country{Germany}
}
\author{Paul Prasse}
\email{paul.prasse@uni-potsdam.de}
\orcid{0000-0003-1842-3645}
\affiliation{%
  \institution{University of Potsdam}
  \city{Potsdam}
  \country{Germany}
}
\author{David R. Reich}
\orcid{0000-0002-3524-3788}
\email{david.reich@uni-potsdam.de}
\affiliation{%
  \institution{University of Potsdam}
  \city{Potsdam}
  \country{Germany}
}
\author{Sebastian Lapuschkin}
\orcid{0000-0002-0762-7258}
\email{sebastian.lapuschkin@hhi.fraunhofer.de}
\affiliation{%
  \institution{Fraunhofer Heinrich Hertz Institute}
  \city{Berlin}
  \country{Germany}
}
\author{Tobias Scheffer}
\email{tobias.scheffer@uni-potsdam.de}
\orcid{0000-0003-4405-7925}
\affiliation{%
  \institution{University of Potsdam}
  \city{Potsdam}
  \country{Germany}
}
\author{Lena A. J\"{a}ger}
\email{jaeger@cl.uzh.ch}
\orcid{0000-0001-9018-9713}
\affiliation{%
  \institution{University of Zurich}
  \city{Zurich}
  \country{Switzerland}\\
  \institution{University of Potsdam}
  \city{Potsdam}
  \country{Germany}
}

\begin{abstract}
    Recent work in XAI for eye tracking data has evaluated the suitability of feature attribution methods to explain the output of deep neural sequence models for the task of oculomotric biometric identification. These methods provide saliency maps to highlight important input features of a specific eye gaze sequence. However, to date, its localization analysis has been lacking a quantitative approach across entire datasets. In this work, we employ established gaze event detection algorithms for fixations and saccades and quantitatively evaluate the impact of these events by determining their \textit{concept influence}. Input features that belong to saccades are shown to be substantially more important than features that belong to fixations. By dissecting saccade events into sub-events, we are able to show that gaze samples that are close to the saccadic peak velocity are most influential. We further investigate the effect of event properties like saccadic amplitude or fixational dispersion on the resulting concept influence.
\end{abstract}




\maketitle

\begin{figure*}[t!]
\centering
\includegraphics[width=0.95\linewidth]{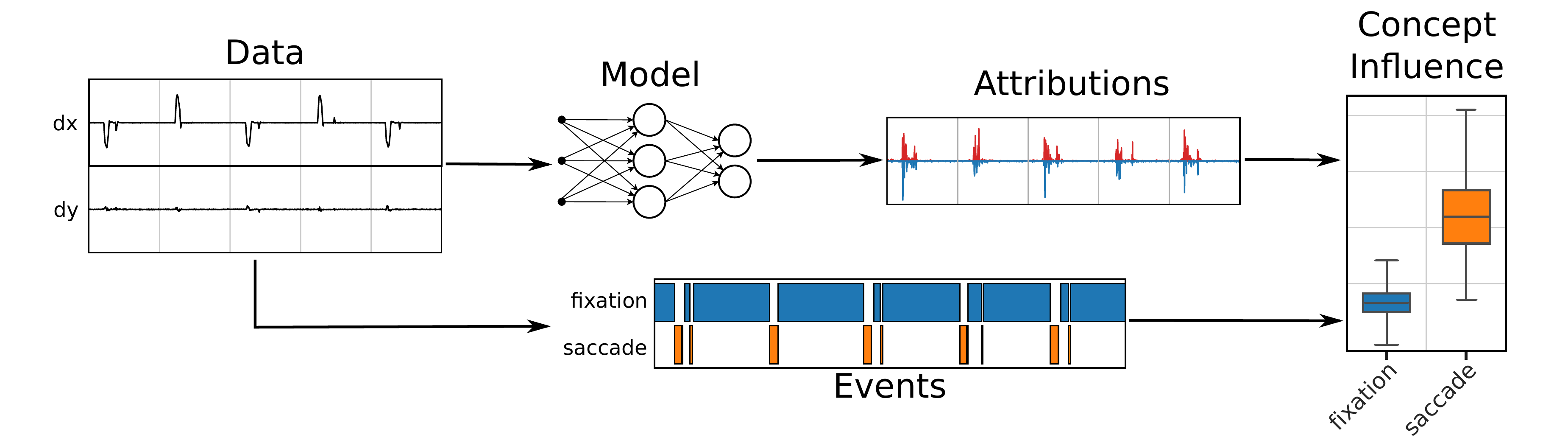}
\caption{The overall process for evaluating the concept influences of distinct types of gaze events.}
\label{fig:methodology}
\end{figure*}

\section{Introduction \& Related Work}

Deep neural networks led to considerable advances in the domains of computer vision~\cite{Voulodimos2018}, speech recognition~\cite{Nassif2019}, time series analysis~\cite{IsmailFawaz2019} and gaze analysis~\cite{Jaeger2019}.
It has been widely observed that end-to-end training of deep neural networks, that is, using unprocessed data as input to the network and let the model learn internal representations, typically outperforms approaches that use aggregated data and engineered features as model input~\cite{Krizhevsky2012}. 
We observe this trend also in gaze analysis, where several network architectures have been presented over the last years that improved on the state-of-the-art by using the non-aggregated gaze velocity time series as input. 
Breakthroughs were especially made on the task of oculomotoric biometric identification, where large improvements in performance were observed that even went along with a decrease in the required duration of input sequence length for successful identification~\cite{Jaeger2019,Makowski2021,Lohr2021,Lohr2022}.

The downside of these complex neural networks is their black box nature as they are generally not interpretable.
This issue is particularly acute for the vast amount of potential medical applications of gaze analysis, such as the detection of autism~\cite{Alcaniz2021,jiang2017learning}, ADHD~\cite{Deng2022} or developmental language disorders~\cite{Key2020,raatikainen2021detection}, as medical applications typically require explainable model predictions.

To provide explanations alongside model predictions, local post-hoc feature attribution methods have been developed to provide saliency maps of the relevance of each input feature~\cite{Shrikumar2016,Sundararajan2017,Bach2015}.
This way we can visualize the positive and negative impact of the input and point to the most important parts of the input that led to a specific model decision.
These methods are local in the sense that they are computed for a single data instance, and they are post-hoc as they are applied to an existing trained model~\cite{Molnar2022}.

However, in the context of end-to-end trained neural networks without explicit feature engineering we also have an input space that is much less interpretable than engineered features.
That leads to \textit{feature attribution methods} coming short in interpretability as saliency maps lack semantic concepts in complex input spaces.
Moreover, global insights across whole datasets are desired to explain the intertwined system of data and trained model.
To overcome this limitation, we can evaluate the impact of the occurence of particular semantic concepts in the input signal, a method named \emph{concept influence}~\cite{Theiner2022}.

Alternative methods to evaluate the impact of semantic concepts on the model output include the inpainting or masking of specific concepts and quantifying the change of output of the neural network~\cite{Williford2020}.
Moreover, internal representations of neural networks can be harnessed to generate concept activation vectors~\cite{Kim2017}.

Gaze events like fixations and saccades are main descriptive concepts of eye tracking research~\cite{Holmqvist2011} and thus are natural candidates for semantic concepts in gaze signals.
By evaluating the \emph{concept influence} of these gaze concepts across entire datasets we generate insights on both model and data.
Prior eye gaze related research either uses descriptive input features on machine learning models~\cite{rigas2016biometric} or is limited to statistical analysis~\cite{rigas2018study,Holland2013} to gain insights on the impact of semantic concepts.
First studies on the interpretability of models in the eye tracking domain are carried out by~\citet{Kumar2020} and feature a PCA analysis of the embedding layer of a neural network.
Further, \citet{Krakowczyk2022} evaluate several feature attribution methods which are applied in the context of oculomotoric biometric identification.

This work in turn puts forward the following contributions for evaluating deep neural sequence models:
\begin{itemize}
    \item Evaluation of \emph{concept influence} of the basic event types saccades and fixations;
    \item Dissection of saccades into sub-events and evaluation of their \emph{concept influences};
    \item Investigation of the relationship between event properties and the resulting \emph{concept influence}.
\end{itemize}


\section{Problem Setting}
\label{sec:problem-setting}

We investigate the explainability of models which get eye gaze velocity time-series data as input and which are trained in a biometric identification setting.
We choose this specific task for its most striking performance benefits over traditional feature engineering approaches.

Given eye gaze time-series data $X \subset \mathbb{R}^{N \times D \times L}$ with $N$ instances, $D$ channels and a 
sequence length of $L$, and a one-hot coded participant labeling $Y \in \{0, 1\}^{N \times K}$ with $K$ labels, we can train a biometric model~$\phi:~\mathbb{R}^{D \times L}~\rightarrow~\mathbb{R}^{K}$ in both a multiclass and a metric learning setting to output the presumed identity of a recorded participant~\cite{Lohr2022}. We can further create a local post-hoc feature attribution function $f(\mathbf{x}, \phi)$ which attributes a relevance to each input feature of any instance $\mathbf{x}$ in regard to the output of the actual model $\phi$~\cite{Krakowczyk2022}.

For image-data these feature attributions are often called pixel-wise explanations~\cite{Bach2015}, and although referred to as \textit{explanations}, they can nevertheless lack interpretability as single pixels are not inherently interpretable by default~\cite{Theiner2022}. This problem can intensify with multi-channel time-series data, as the input space is potentially visually less interpretable. 

By measuring the overlap of the highest attributed input features and an interpretable segmentation that refers to a specific \textit{concept}, we can compute the \textit{concept influence} of this concept~\cite{Theiner2022}. Whereas in the image-domain, segmentations are sets of pixels that are associated with detected objects like houses, streets, persons or the sky, in the time-series domain, segmentations refer to events delimited by their on- and offsets. In the case of eye tracking data, gaze events suggest themselves as interpretable concepts for our study, as their function and underlying neuro-biological processes have been extensively researched over the past decades~\cite{Rayner1983, Rayner2004, Martinez-Conde2004, Martinez-Conde2006, EngbertKliegl2003}, and there also exist a vast amount of methods to automatically detect them~\cite{Andersson2016,Startsev2022}.

When investigating which parts of the input sequence are most relevant for the output of deep neural sequence models, we can make use of the established concepts of distinct gaze events and evaluate which ones of these exhibit the highest influence on the model output.

\section{Materials and Methods}

This section is structured alongside Figure~\ref{fig:methodology}, where we illustrate the overall evaluation process.
We start out by presenting the used datasets in Subsection~\ref{sec:data-sets} and subsequently describe our data preprocessing steps in Subsection~\ref{sec:data-preprocessing}.
We continue by introducing the employed algorithms for gaze event detection in Subsection~\ref{sec:event-detection} and detail the method for dissecting a saccadic event into sub-events in Subsection~\ref{sec:event-dissection}.
Subsection~\ref{sec:biometric-model} briefly covers the biometric task and model under investigation, whereas Subsection~\ref{sec:attribution-methods} gives an overview of the employed attribution methods.
We delineate the term \emph{concept influence} in Subsection~\ref{sec:concept-influence}.
The overall evaluation protocol is detailed in Subsection~\ref{sec:evaluation-protocol}.

\subsection{Dataset}
\label{sec:data-sets}

We make use of the three publicly available datasets \emph{GazeBase}~\cite{Griffith2021}, \emph{JuDo1000}~\cite{Makowski2020b} and the \emph{Potsdam Textbook Corpus (PoTeC)}~\cite{Jaeger2021b}.
All datasets are recorded at a sampling rate of~\SI{1000}{\Hz}. Only \emph{JuDo1000} contains binocular recordings. We have reduced the GazeBase dataset to the first 4~rounds where most subjects participated in. We use all of the available stimuli for evaluation.
Table~\ref{tab:datasets} in Appendix~\ref{apx:datasets} provides a brief summary of dataset properties.

\subsection{Data Preprocessing}
\label{sec:data-preprocessing}

Based on the preprocessing pipeline of~\citet{Lohr2022}, we apply the Savitzky-Golay differentiation filter~\cite{SavitzkyGolay1964} with a window size of 7 and an order of 2 to transform positional data into gaze velocity data.
We construct subsequences by a non-overlapping rolling window with a window size of~\SI{5}{\second} for \emph{GazeBase} and~\SI{1}{\second} for \emph{JuDo1000} and \emph{PoTeC}.
We clamp velocities to~\SI{\pm1000}{\degree/\second} and exclude all subsequences which would need padding or comprise more than 50\% missing values.
We finally apply z-score normalization and replace all missing values with 0.
We leverage the pymovements package for these preprocessing steps~\cite{pymovements}.

\subsection{Gaze Event Detection Algorithms}
\label{sec:event-detection}

We limit our study to the investigation of fixations and saccades as these two event types are the main ones investigated in the literature.
Further candidates would have been post-saccadic oscillations, smooth pursuit and blinks, but also micromovements like drift and tremor.
Note that we do not distinguish between saccades and microsaccades in this work and include microsaccades in the set of saccades.
For binocular data we solely use the right eye for event detection.

We use distinct detection algorithms for fixations and saccades as suggested by Andersson~et~al.~\cite{Andersson2016}.
We use the I-VT algorithm~\cite{Salvucci2000} to detect fixations and the algorithm of \citet{EngbertKliegl2003} to detect saccades.
Table~\ref{tab:event-detection-parameters} in Appendix~\ref{apx:event-detection-parameters} lists all parameters used in the event detection process.
Although the I-VT algorithm originally just uses a single parameter for its fixation velocity threshold, we make use of an additional minimum fixation duration and maximum fixation dispersion threshold to avoid misclassifications. Fixations that exceed these values will be simply excluded from evaluation.

The main parameter of the employed saccade detection algorithm is the threshold factor which is multiplied with the adaptively determined noise threshold.
We further exclude saccades from evaluation which exceed the valid ranges for saccade duration and peak velocity stated in Table~\ref{tab:event-detection-parameters} in Appendix~\ref{apx:event-detection-parameters}.

\begin{figure}
\centering
\includegraphics[width=\linewidth]{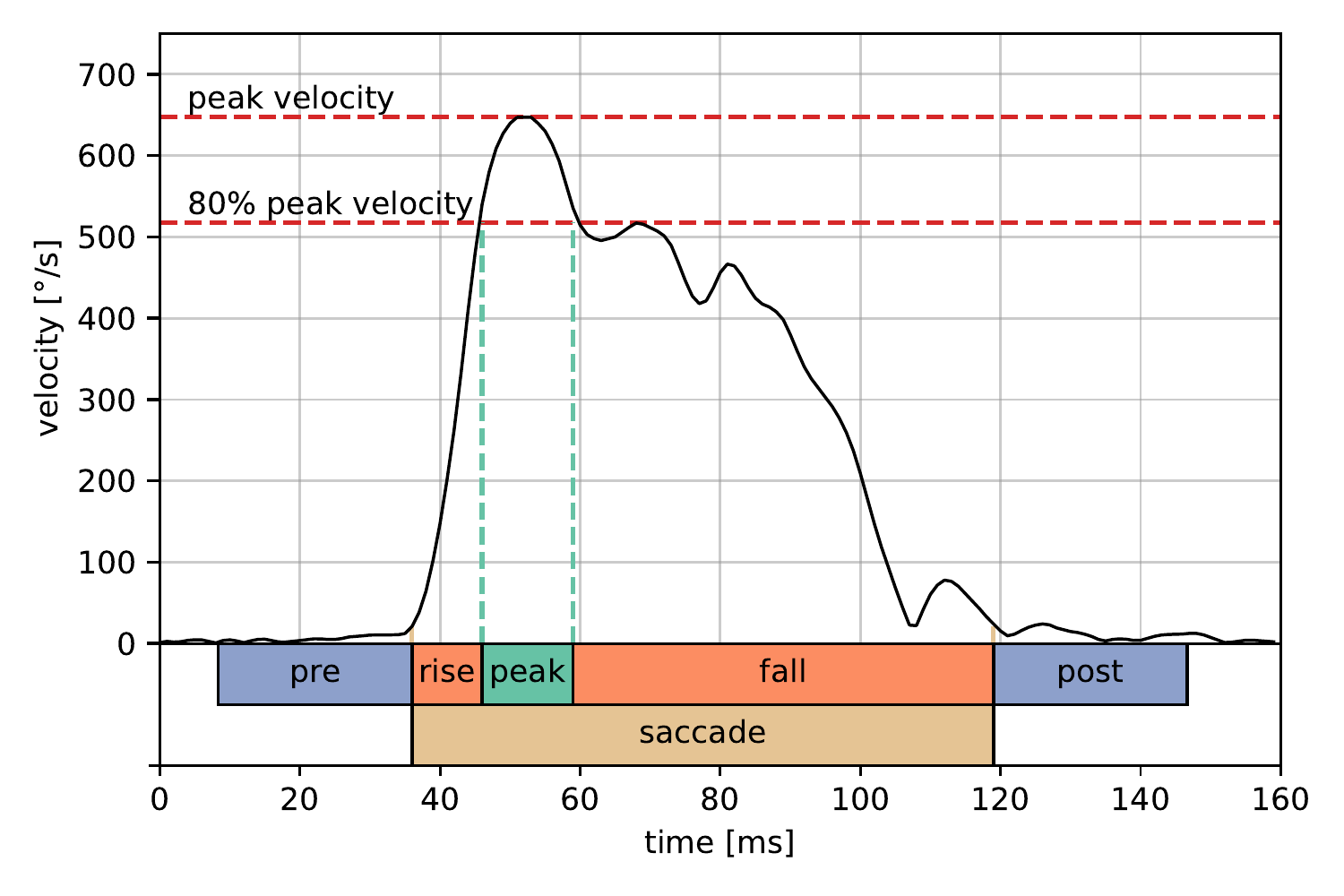}
\caption{Illustration of the saccade event dissection algorithm. The saccade event, depicted as the brown horizontal bar at the bottom, is detected by the saccade detection algorithm~\cite{EngbertKliegl2003}.
All samples in a saccade that reach at least 80\% of its peak velocity are associated with the peak phase~(green bar). The rise phase lasts from saccade onset to peak onset, the fall phase lasts from peak offset to saccade offset~(orange bars). Finally we add a pre and post phase at the beginning and the end of a saccade with a duration of 1/3 of the total saccade duration~(blue bars).}
\label{fig:event-dissection-example}
\end{figure}

\subsection{Event Dissection for Saccades}
\label{sec:event-dissection}

Looking at a typical velocity profile of saccadic eye movements, we can identify distinct phases which are illustrated in Figure~\ref{fig:event-dissection-example}.
We see a rapid increase in velocity at the beginning and a velocity decline at the end of a saccade, with a short phase in-between where the gaze velocity is near saccade peak velocity.
We call these sub-events the rise, fall and peak phases. The peak phase is defined as the time steps where the velocity is at least 80\% of the peak velocity of the respective saccade event.
We further chain an additional event to each before the beginning (pre-phase) and after the end (post-phase) of a saccade, with a duration set to be 1/3 of the associated saccade. 
Due to the neglible occurence of samples which are below 80\% of the peak velocity but are between two local peaks greater than 80\% we disregard these samples in our analysis.
This way we get a total of 5 granular events out of a single saccade event.

\subsection{Biometric Model}
\label{sec:biometric-model}

We investigate the explainability of \emph{EyeKnowYouToo}, a state-of-the-art neural network model for oculomotoric biometric identification developed by~\citet{Lohr2022}.
This model is a convolutional network that uses multi-channel sequences of yaw (horizontal) and pitch (vertical) angular gaze velocities as input and that is end-to-end trained to minimize a weighted sum of categorical cross-entropy and multi-similarity loss.
Instead of using the output of the embedding layer for comparison with existing embeddings of an enrollment database as in the original biometric system~\cite{Lohr2022}, we make use of the nodes of the classification layer as targets for calculating attributions of each inference.
We modified Dillon Lohr's model implementation~\cite{Lohr2022} in order to facilitate our application of attribution methods. 
Changes relate to the naming and grouping and layers while the overall model architecture and behavior during training is left the same as in the original.

We restrict our study to this single model as it exhibits the best performance on the given biometric task while also being smallest in the number of model parameters.
Although a comparison of explainability metrics between state-of-the-art biometric models is interesting, such a comparison unfortunately cannot be in scope of our study.

\subsection{Attribution Methods}
\label{sec:attribution-methods}

Feature attribution methods attribute relevance to each input feature, such that saliency maps can be generated to visualize the positive and negative impact of the input for a specific model prediction~\cite{Shrikumar2016,Sundararajan2017,Bach2015}.

Based on the findings of~\citet{Krakowczyk2022}, we limit this study to the three best performing methods: DeepLIFT~(DL)~\cite{Shrikumar2017}, Integrated Gradients~(IG)~\cite{Sundararajan2017} and Layer-wise Relevance Propagation~(LRP)~\cite{Bach2015,Montavon2017,Montavon2018}. We use the Zennit library~\citep{Anders2021} for the implementation of LRP rules and the Captum library~\citep{Kokhlikyan2020} for its DeepLIFT and IG implementations.

All three methods are backpropagation-based in the way that they propagate the relevance of the model output back to each input feature~\cite{Ancona2017}.
DeepLIFT and IG additionally require a baseline reference input which is desired to generate neutral model output and supposed to have low relevance across all input features.
We set this baseline to zero in concordance with predominant usage~\cite{Sturmfels2020}.

IG attributes relevance by computing the gradients of a model with respect to each input feature. Input features are step-wise linearly interpolated from the reference baseline into the given input instance. The integral of the gradients along this interpolatation path is multiplied by the difference between reference baseline and given input instance~\cite{Sundararajan2017}.

LRP attributions are computed by backpropagating the model output layer by layer. Depending on the product of activations and weights of the incoming connections, relevance of each unit is passed down to the preceding layer.
We limit this study to the vanilla LRP-$\varepsilon$~rule~\cite{Kohlbrenner2020} and set $\varepsilon = 0.25$~\cite{Montavon2019}.

DeepLIFT~\cite{Shrikumar2017} is similar to the former layer-wise backpropagation method but uses the reference baseline to calculate activation reference points for each unit. Activation differences to the reference points are then backpropagated as relevance.

\subsection{Concept Influence}
\label{sec:concept-influence}

As stated in the problem setting in Section~\ref{sec:problem-setting}, the drawback of pixel-wise feature attributions is that pixels are usually not inherently interpretable on their own.
When we ask which parts of the input have the biggest impact on the output of the model under investigation, we usually expect the explanation to be given in interpretable high-level concepts instead of a simple saliency map.
We further do not want to be limited to local attributions computed for individual data instances only, but aim at global inferences about the dataset and model as a whole.

The concept influence method proposed by~\citet{Theiner2022} which is originally developed for image data, tackles this issue by quantifying the overlap between given concepts and the highest attributions.
To this end, each concept is represented as a binary segmentation $S \in \{0, 1\}^{L}$, where $S_i = 1$ encodes the presence of the respective concept at the specific step~$i$ in the sequence of length~$L$.
We refer to the segmentation $S$ as the \textit{concept segmentation}, with its size~$|S|$ being defined as $\sum_{i=1}^{L}{S_i}$ given the sequence length~$L$.

We further create a second segmentation $T \in \{0, 1\}^{L}$ from the top-$k$ highest feature attribution values, after squashing multi-channel feature attributions to a single channel by taking the step-wise maximum.
As in the original work by~\citet{Theiner2022} we set $k$ to be~\SI{2}{\%}~of the input size (\num{20} for an $L$ of \num{1000}, \num{100} for an $L$ of \num{5000}).

To measure the influence of a specific concept, we take the size of the intersection of the concept segmentation $S$ and the top-$k$ attribution segmentation $T$.
This intermediate result is called the top-$k$~intersection~\cite{Theiner2022}.
To account for the fact that the size of the concept segmentation has an impact on the resulting intersection, we perform a weighting by dividing the top-k~intersection by the size of the segmentation~$|S|$ relative to the sequence length~$L$ as defined in Equation~\ref{eq:ci}.

\begin{equation}
    \text{c} = \frac{L}{|S|} ~~ \frac{1}{k} ~~ \sum_{i=1}^{L}{S_i \land T_i}
  \label{eq:ci}
\end{equation}

The resulting value $c$ is the \emph{concept influence} for the respective concept and ranges between 0 and ${L}\;/\;{|S|}$. A concept influence above 1 is regarded as highly influential~\cite{Theiner2022}.

\subsection{Evaluation Protocol}
\label{sec:evaluation-protocol}

We train model weights and subsequently generate attributions by applying a k-fold cross-validation protocol which splits data into training and test sets.
We use different schemes for each dataset to split data into folds: leave-one-round-out for \emph{GazeBase}~($k = 4$), leave-one-session-out for \emph{JuDo1000}~($k = 4$), and leave-one-text-out for \emph{PoTeC}~($k = 12$).
We compute attributions on the test set and set the predicted class as the target class for relevance computation.

We implement the general evaluation framework using scikit-learn~\citep{scikit-learn} und use the attribution metric implementations from Quantus~\cite{Hedstrom2023}.
Our hardware setup comprises an AMD EPYC 7742 CPU and a NVIDIA DGX A100 GPU.
The code can be found online.\footnote{https://github.com/aeye-lab/etra-2023-bridging-the-gap}

\section{Experimental Results}

We present our experimental results in the following section. All attributions are evaluated on the model described in Section~\ref{sec:biometric-model}.
Figure~\ref{fig:model-accuracy} in Appendix~\ref{apx:model-accuracy} reports accuracies of at least 90\% for each used dataset.
We begin by putting forward the concept influence evaluation of the basic event types in Section~\ref{sec:results-basic-events}, whereas Section~\ref{sec:results-event-dissection} deals with the concept influences of saccade sub-events.
We bin fixations and saccades according to several properties and show our results in Section~\ref{sec:results-property-binning}.

\subsection{Concept Influence of Fixations and Saccades}
\label{sec:results-basic-events}


\begin{figure}
\centering
\includegraphics[width=0.9\linewidth]{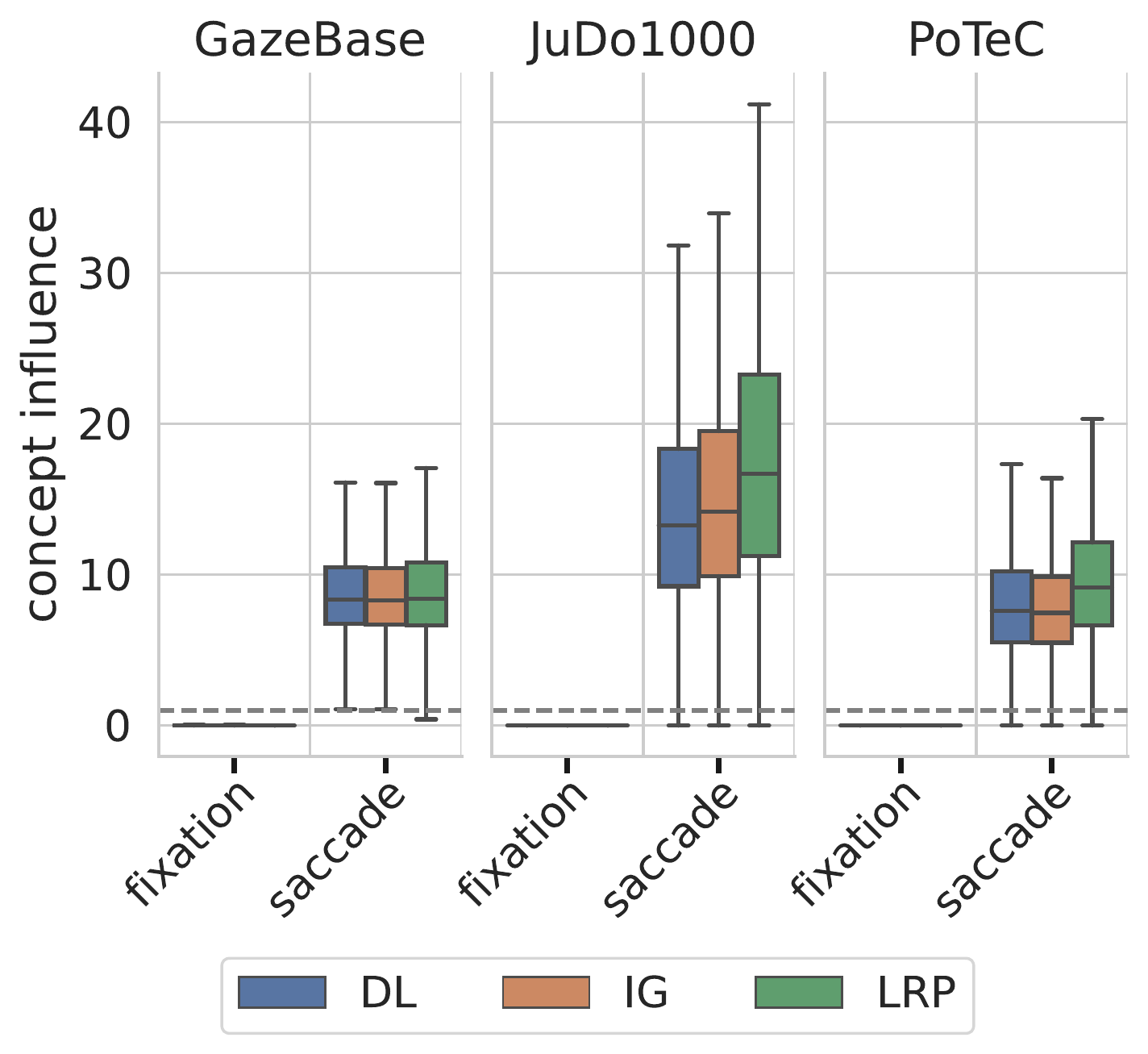}
\caption{Concept influences for saccades and fixations}
\label{fig:event-type-ci}
\end{figure}

Out of the detected saccades and fixations from Section~\ref{sec:event-detection} we create concept segmentations for each event type.
The distribution of segmentation sizes is depicted in Figure~\ref{fig:event-type-sizes} in Appendix~\ref{apx:event-type-sizes}.
Across all three datasets we observe a much greater segmentation size for fixations than for saccades.

The resulting concept influences in Figure~\ref{fig:event-type-ci} demonstrate very high values for saccades instead.
In contrast, fixations rarely exceed a concept influence of 0.1 which assesses their influence on model prediction to be very limited. This holds true across all datasets and attribution methods.
We can conclude that according to the chosen evaluation method, saccades have a much bigger concept influence and thus their velocity profiles contain more information with respect to the problem setting of biometric identification than it is the case for fixations.

\subsection{Saccade Sub-Event Dissection}
\label{sec:results-event-dissection}

\begin{figure}
\centering
\includegraphics[width=\linewidth]{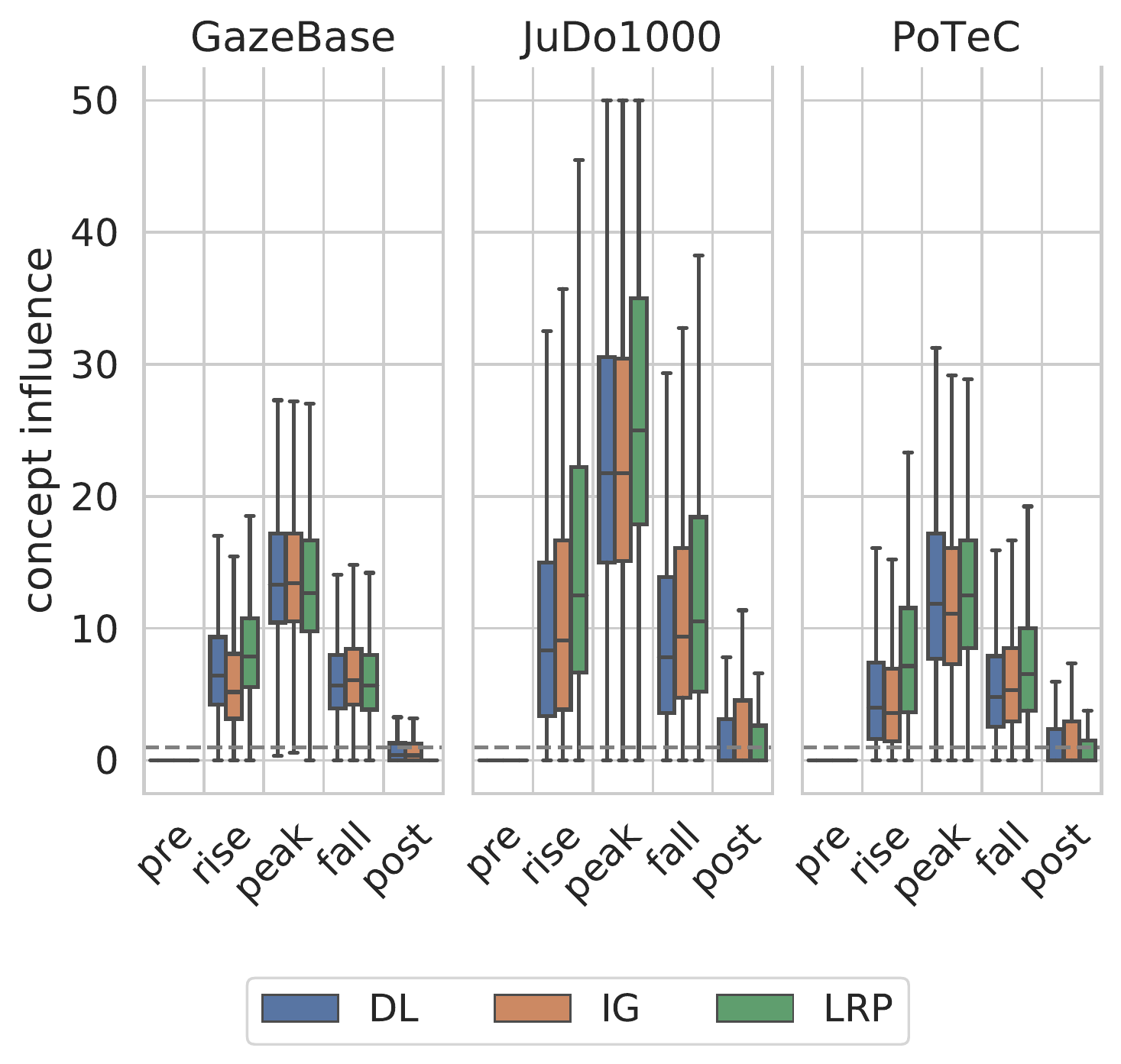}
\caption{Concept influences for saccade sub-events.}
\label{fig:event-dissection-ci}
\end{figure}

We further present the results for the event dissection experiment where we dissect saccades into sub-events as shown in Figure~\ref{fig:event-dissection-ci}.
The distribution of the sub-event segmentation sizes can be found in Figure~\ref{fig:event-dissection-sizes} in Appendix~\ref{apx:event-dissection-sizes}.
Across all datasets and attribution methods, we observe the highest concept influence for samples belonging to the peak phase of the saccadic profile.
The concept influences of rise and fall phases are both about half as high as the peak phase.
We note close to no concept influence for the pre phase, but a moderately influential post phase which can be associated with the occurence of post-saccadic oscillations.



\subsection{Event Property Binning}
\label{sec:results-property-binning}

\begin{figure*}
\centering
\includegraphics[width=0.3\linewidth]{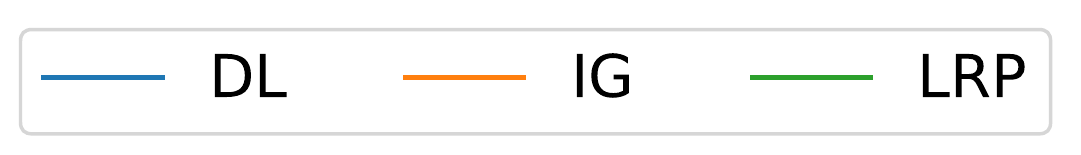}
\subfigure[Property binning by saccade duration.]{\includegraphics[width=0.99\linewidth]{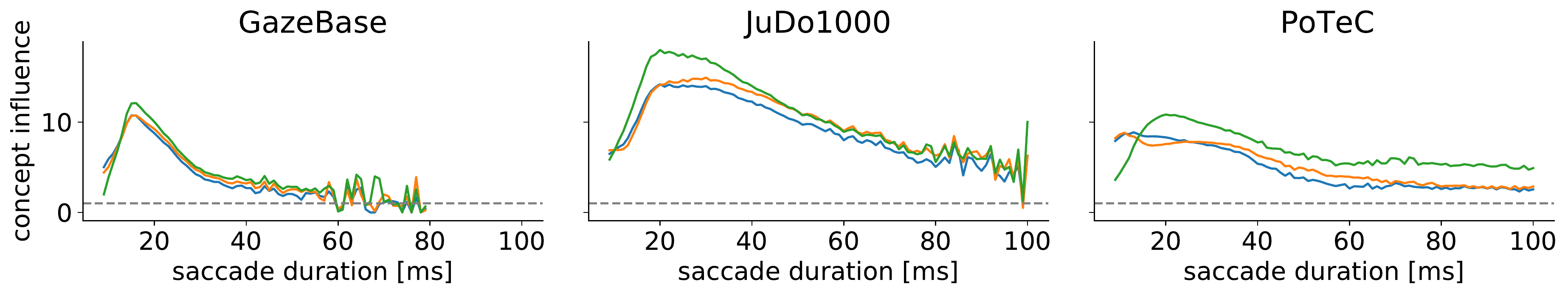}\label{fig:event-property-binning-saccade-duration-ci}}\\
\subfigure[Property binning by saccade amplitude.]{\includegraphics[width=0.99\linewidth]{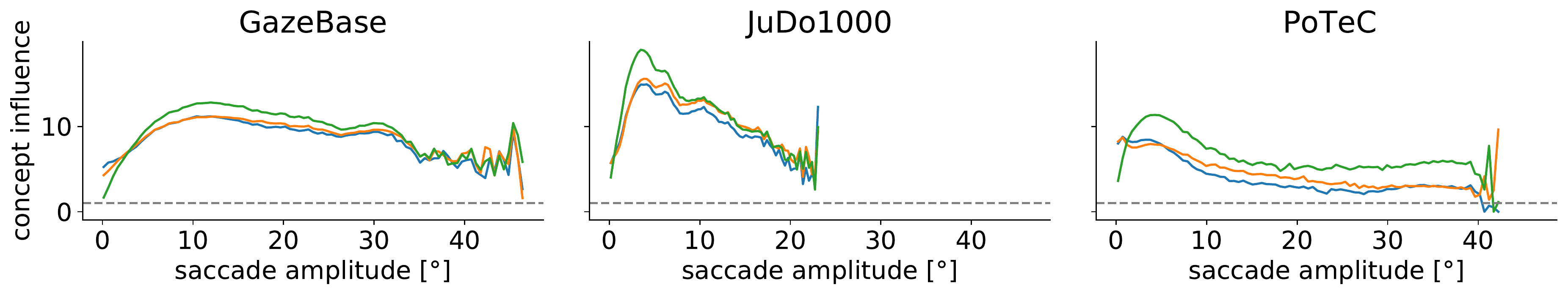}\label{fig:event-property-binning-saccade-amplitude-ci}}\\
\subfigure[Property binning by fixation dispersion.]{\includegraphics[width=0.99\linewidth]{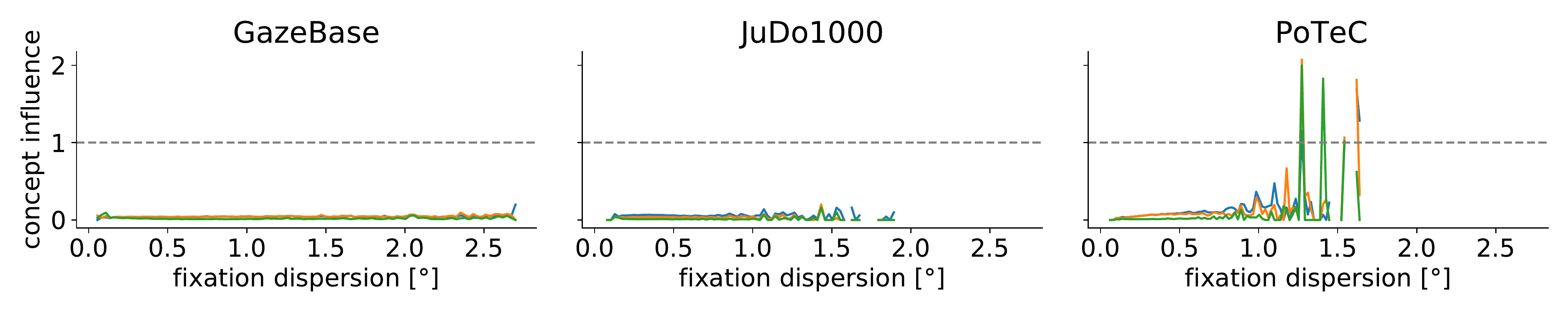}\label{fig:event-property-binning-fixation-dispersion-ci}}\\
\subfigure[Property binning by fixation velocity standard deviation.]{\includegraphics[width=0.99\linewidth]{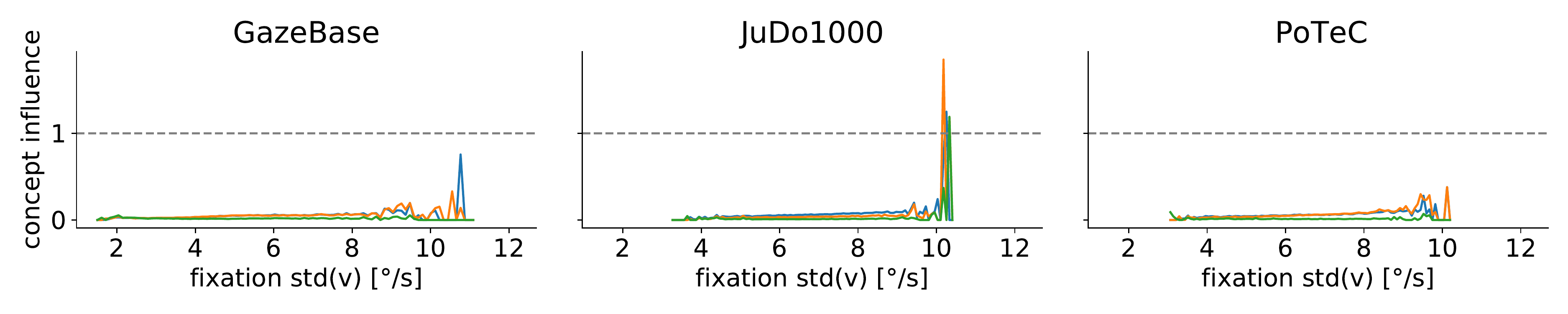}\label{fig:event-property-binning-fixation-vstd-ci}}\\
\caption{Results for the event property binning experiment. The dashed gray horizontal line represents a concept influence of 1.}
\label{fig:event-property-binning-saccade-amplitude}
\end{figure*}

In this subsection we study the effect of different event properties on the resulting concept influence.
We select two properties for each event type: we study the duration and amplitude of saccades and the dispersion and velocity standard deviation of fixations.
The distribution of the segmentation sizes across these properties can be found in Figure~\ref{fig:event-property-binning-saccade-amplitude} in Appendix~\ref{apx:event-property-binning-saccade-amplitude}.

Starting with the duration of saccades in Figure~\ref{fig:event-property-binning-saccade-duration-ci}, we observe the highest concept influences for saccades with a duration of about \SI{20}{\ms}.
We further observe a concept influence peak for saccade amplitudes below \SI{10}{\degree}  on the JuDo1000 and PoTeC datasets in Figure~\ref{fig:event-property-binning-saccade-amplitude-ci}. Regarding the GazeBase dataset the peak is much less pronounced and is slightly higher at about \SI{10}{\degree}.
Continuing with properties of fixation events, we mostly see flat curves and depending on the data set we spot rare outliers on the upper bounds.
In the case of fixation dispersion in Figure~\ref{fig:event-property-binning-fixation-dispersion-ci}, we observe concept influences above 1 solely on high dispersion outliers of the PoTeC dataset.
In the case of the standard deviation of velocities during fixations presented in Figure~\ref{fig:event-property-binning-fixation-vstd-ci}, we observe a rise in concept influence on high standard deviations, but concept influences above 1 are solely reached on the JuDo1000 dataset.

\section{Discussion \& Conclusion}
\label{sec:discussion}

We have demonstrated the feasibility of evaluating the \emph{concept influence} of gaze event types to gain insights on which parts of a gaze sequence is most relevant for the classification process of a state-of-the-art biometric model.
We observed high concept influences for saccades with the peak phase of a saccade event to be especially influential.
In contrast, fixations exhibit neglible concept influences with the exception of fixations with a high dispersion or a high standard deviation in velocity during fixation.

Although the specific results of this study are very much constrained to the oculomotoric biometric setting, this work serves as a frame work for further research on the explainability of deep neural sequence models that consume gaze time-series data. This way we can harness the best of both worlds: top performance from neural networks and interpretable insights from descriptive concepts.

\begin{acks}
This work was partially funded by the German Federal Ministry of Education and Research under grant 01IS20043.
\end{acks}

\bibliographystyle{ACM-Reference-Format}
\bibliography{bibliography}


\begin{thebibliography}{52}


\ifx \showCODEN    \undefined \def \showCODEN     #1{\unskip}     \fi
\ifx \showDOI      \undefined \def \showDOI       #1{#1}\fi
\ifx \showISBNx    \undefined \def \showISBNx     #1{\unskip}     \fi
\ifx \showISBNxiii \undefined \def \showISBNxiii  #1{\unskip}     \fi
\ifx \showISSN     \undefined \def \showISSN      #1{\unskip}     \fi
\ifx \showLCCN     \undefined \def \showLCCN      #1{\unskip}     \fi
\ifx \shownote     \undefined \def \shownote      #1{#1}          \fi
\ifx \showarticletitle \undefined \def \showarticletitle #1{#1}   \fi
\ifx \showURL      \undefined \def \showURL       {\relax}        \fi
\providecommand\bibfield[2]{#2}
\providecommand\bibinfo[2]{#2}
\providecommand\natexlab[1]{#1}
\providecommand\showeprint[2][]{arXiv:#2}

\bibitem[\protect\citeauthoryear{Alca{\~{n}}iz, Chicchi-Giglioli,
  Carrasco-Ribelles, Mar{\'{\i}}n-Morales, Minissi, Teruel-Garc{\'{\i}}a,
  Sirera, and Abad}{Alca{\~{n}}iz et~al\mbox{.}}{2021}]%
        {Alcaniz2021}
\bibfield{author}{\bibinfo{person}{Mariano Alca{\~{n}}iz},
  \bibinfo{person}{Irene~Alice Chicchi-Giglioli},
  \bibinfo{person}{Luc{\'{\i}}a~A. Carrasco-Ribelles}, \bibinfo{person}{Javier
  Mar{\'{\i}}n-Morales}, \bibinfo{person}{Maria~Eleonora Minissi},
  \bibinfo{person}{Gonzalo Teruel-Garc{\'{\i}}a}, \bibinfo{person}{Marian
  Sirera}, {and} \bibinfo{person}{Luis Abad}.} \bibinfo{year}{2021}\natexlab{}.
\newblock \showarticletitle{Eye gaze as a biomarker in the recognition of
  autism spectrum disorder using virtual reality and machine learning: A proof
  of concept for diagnosis}.
\newblock \bibinfo{journal}{\emph{Autism Research}} \bibinfo{volume}{15},
  \bibinfo{number}{1} (\bibinfo{date}{nov} \bibinfo{year}{2021}),
  \bibinfo{pages}{131--145}.
\newblock
\urldef\tempurl%
\url{https://doi.org/10.1002/aur.2636}
\showDOI{\tempurl}


\bibitem[\protect\citeauthoryear{Ancona, Ceolini, Öztireli, and Gross}{Ancona
  et~al\mbox{.}}{2017}]%
        {Ancona2017}
\bibfield{author}{\bibinfo{person}{Marco Ancona}, \bibinfo{person}{Enea
  Ceolini}, \bibinfo{person}{Cengiz Öztireli}, {and} \bibinfo{person}{Markus
  Gross}.} \bibinfo{year}{2017}\natexlab{}.
\newblock \showarticletitle{Towards better understanding of gradient-based
  attribution methods for Deep Neural Networks}.
\newblock \bibinfo{journal}{\emph{arXiv preprint arXiv:1711.06104}}
  (\bibinfo{year}{2017}).
\newblock


\bibitem[\protect\citeauthoryear{Anders, Neumann, Samek, Müller, and
  Lapuschkin}{Anders et~al\mbox{.}}{2021}]%
        {Anders2021}
\bibfield{author}{\bibinfo{person}{Christopher~J. Anders},
  \bibinfo{person}{David Neumann}, \bibinfo{person}{Wojciech Samek},
  \bibinfo{person}{Klaus-Robert Müller}, {and} \bibinfo{person}{Sebastian
  Lapuschkin}.} \bibinfo{year}{2021}\natexlab{}.
\newblock \showarticletitle{Software for Dataset-wide {XAI}: From Local
  Explanations to Global Insights with {Zennit}, {CoRelAy}, and {ViRelAy}}.
\newblock \bibinfo{journal}{\emph{arXiv preprint arXiv:2106.13200}}
  (\bibinfo{year}{2021}).
\newblock


\bibitem[\protect\citeauthoryear{Andersson, Larsson, Holmqvist, Stridh, and
  Nyström}{Andersson et~al\mbox{.}}{2016}]%
        {Andersson2016}
\bibfield{author}{\bibinfo{person}{Richard Andersson}, \bibinfo{person}{Linnea
  Larsson}, \bibinfo{person}{Kenneth Holmqvist}, \bibinfo{person}{Martin
  Stridh}, {and} \bibinfo{person}{Marcus Nyström}.}
  \bibinfo{year}{2016}\natexlab{}.
\newblock \showarticletitle{One algorithm to rule them all? An evaluation and
  discussion of ten eye movement event-detection algorithms}.
\newblock \bibinfo{journal}{\emph{Behavior Research Methods}}
  \bibinfo{volume}{49}, \bibinfo{number}{2} (\bibinfo{year}{2016}),
  \bibinfo{pages}{616--637}.
\newblock


\bibitem[\protect\citeauthoryear{Bach, Binder, Montavon, Klauschen, Müller,
  and Samek}{Bach et~al\mbox{.}}{2015}]%
        {Bach2015}
\bibfield{author}{\bibinfo{person}{Sebastian Bach}, \bibinfo{person}{Alexander
  Binder}, \bibinfo{person}{Grégoire Montavon}, \bibinfo{person}{Frederick
  Klauschen}, \bibinfo{person}{Klaus-Robert Müller}, {and}
  \bibinfo{person}{Wojciech Samek}.} \bibinfo{year}{2015}\natexlab{}.
\newblock \showarticletitle{On Pixel-Wise Explanations for Non-Linear
  Classifier Decisions by Layer-Wise Relevance Propagation}.
\newblock \bibinfo{journal}{\emph{PLOS ONE}} \bibinfo{volume}{10},
  \bibinfo{number}{7} (\bibinfo{year}{2015}), \bibinfo{pages}{1--46}.
\newblock


\bibitem[\protect\citeauthoryear{Deng, Prasse, Reich, Dziemian,
  Stegenwallner-Sch\"{u}tz, Krakowczyk, Makowski, Langer, Scheffer, and
  J\"{a}ger}{Deng et~al\mbox{.}}{2023}]%
        {Deng2022}
\bibfield{author}{\bibinfo{person}{Shuwen Deng}, \bibinfo{person}{Paul Prasse},
  \bibinfo{person}{David~R. Reich}, \bibinfo{person}{Sabine Dziemian},
  \bibinfo{person}{Maja Stegenwallner-Sch\"{u}tz}, \bibinfo{person}{Daniel
  Krakowczyk}, \bibinfo{person}{Silvia Makowski}, \bibinfo{person}{Nicolas
  Langer}, \bibinfo{person}{Tobias Scheffer}, {and} \bibinfo{person}{Lena~A.
  J\"{a}ger}.} \bibinfo{year}{2023}\natexlab{}.
\newblock \showarticletitle{Detection Of {ADHD} Based On Eye Movements During
  Natural Viewing}. In \bibinfo{booktitle}{\emph{Machine Learning and Knowledge
  Discovery in Databases: European Conference, ECML PKDD 2022, Grenoble,
  France, September 19–23, 2022, Proceedings, Part VI}} (Grenoble, France).
  \bibinfo{publisher}{Springer-Verlag}, \bibinfo{address}{Berlin, Heidelberg},
  \bibinfo{pages}{403–418}.
\newblock
\showISBNx{978-3-031-26421-4}
\urldef\tempurl%
\url{https://doi.org/10.1007/978-3-031-26422-1_25}
\showDOI{\tempurl}


\bibitem[\protect\citeauthoryear{Dombrow}{Dombrow}{2018}]%
        {Dombrow2018}
\bibfield{author}{\bibinfo{person}{Isabel Dombrow}.}
  \bibinfo{year}{2018}\natexlab{}.
\newblock \showarticletitle{Saccadic inhibition in a guided saccade task}.
\newblock \bibinfo{journal}{\emph{PeerJ}} (\bibinfo{year}{2018}).
\newblock


\bibitem[\protect\citeauthoryear{Engbert and Kliegl}{Engbert and
  Kliegl}{2003}]%
        {EngbertKliegl2003}
\bibfield{author}{\bibinfo{person}{R. Engbert} {and} \bibinfo{person}{R.
  Kliegl}.} \bibinfo{year}{2003}\natexlab{}.
\newblock \showarticletitle{Microsaccades uncover the orientation of covert
  attention}.
\newblock \bibinfo{journal}{\emph{Vision Research}}  \bibinfo{volume}{43}
  (\bibinfo{year}{2003}), \bibinfo{pages}{1035--1045}.
\newblock


\bibitem[\protect\citeauthoryear{Fawaz, Forestier, Weber, Idoumghar, and
  Muller}{Fawaz et~al\mbox{.}}{2019}]%
        {IsmailFawaz2019}
\bibfield{author}{\bibinfo{person}{Hassan~Ismail Fawaz},
  \bibinfo{person}{Germain Forestier}, \bibinfo{person}{Jonathan Weber},
  \bibinfo{person}{Lhassane Idoumghar}, {and} \bibinfo{person}{Pierre-Alain
  Muller}.} \bibinfo{year}{2019}\natexlab{}.
\newblock \showarticletitle{Deep learning for time series classification: a
  review}.
\newblock \bibinfo{journal}{\emph{Data Mining and Knowledge Discovery}}
  \bibinfo{volume}{33}, \bibinfo{number}{4} (\bibinfo{date}{mar}
  \bibinfo{year}{2019}), \bibinfo{pages}{917--963}.
\newblock
\urldef\tempurl%
\url{https://doi.org/10.1007/s10618-019-00619-1}
\showDOI{\tempurl}


\bibitem[\protect\citeauthoryear{Griffith, Lohr, Abdulin, and
  Komogortsev}{Griffith et~al\mbox{.}}{2021}]%
        {Griffith2021}
\bibfield{author}{\bibinfo{person}{Henry Griffith}, \bibinfo{person}{Dillon
  Lohr}, \bibinfo{person}{Evgeny Abdulin}, {and} \bibinfo{person}{Oleg
  Komogortsev}.} \bibinfo{year}{2021}\natexlab{}.
\newblock \showarticletitle{{GazeBase}, a large-scale, multi-stimulus,
  longitudinal eye movement dataset}.
\newblock \bibinfo{journal}{\emph{Scientific Data}}  \bibinfo{volume}{8}
  (\bibinfo{year}{2021}).
\newblock


\bibitem[\protect\citeauthoryear{Hedström, Weber, Krakowczyk, Bareeva,
  Motzkus, Samek, Lapuschkin, and Höhne}{Hedström et~al\mbox{.}}{2023}]%
        {Hedstrom2023}
\bibfield{author}{\bibinfo{person}{Anna Hedström}, \bibinfo{person}{Leander
  Weber}, \bibinfo{person}{Daniel Krakowczyk}, \bibinfo{person}{Dilyara
  Bareeva}, \bibinfo{person}{Franz Motzkus}, \bibinfo{person}{Wojciech Samek},
  \bibinfo{person}{Sebastian Lapuschkin}, {and} \bibinfo{person}{Marina M.-C.
  Höhne}.} \bibinfo{year}{2023}\natexlab{}.
\newblock \showarticletitle{Quantus: An Explainable AI Toolkit for Responsible
  Evaluation of Neural Network Explanations and Beyond}.
\newblock \bibinfo{journal}{\emph{Journal of Machine Learning Research}}
  \bibinfo{volume}{24}, \bibinfo{number}{34} (\bibinfo{year}{2023}),
  \bibinfo{pages}{1--11}.
\newblock
\urldef\tempurl%
\url{http://jmlr.org/papers/v24/22-0142.html}
\showURL{%
\tempurl}


\bibitem[\protect\citeauthoryear{Holland and Komogortsev}{Holland and
  Komogortsev}{2013}]%
        {Holland2013}
\bibfield{author}{\bibinfo{person}{Corey~D. Holland} {and}
  \bibinfo{person}{Oleg~V. Komogortsev}.} \bibinfo{year}{2013}\natexlab{}.
\newblock \showarticletitle{Complex eye movement pattern biometrics: Analyzing
  fixations and saccades}. In \bibinfo{booktitle}{\emph{2013 International
  Conference on Biometrics (IJCB)}}. \bibinfo{pages}{1--8}.
\newblock


\bibitem[\protect\citeauthoryear{Holmqvist, Nystr{\"o}m, Andersson, Dewhurst,
  Jarodzka, and Van~de Weijer}{Holmqvist et~al\mbox{.}}{2011}]%
        {Holmqvist2011}
\bibfield{author}{\bibinfo{person}{Kenneth Holmqvist}, \bibinfo{person}{Marcus
  Nystr{\"o}m}, \bibinfo{person}{Richard Andersson}, \bibinfo{person}{Richard
  Dewhurst}, \bibinfo{person}{Halszka Jarodzka}, {and} \bibinfo{person}{Joost
  Van~de Weijer}.} \bibinfo{year}{2011}\natexlab{}.
\newblock \bibinfo{booktitle}{\emph{Eye tracking: {A} comprehensive guide to
  methods and measures}}.
\newblock \bibinfo{publisher}{Oxford University Press}.
\newblock


\bibitem[\protect\citeauthoryear{J\"{a}ger, Makowski, Prasse, Liehr, Seidler,
  and Scheffer}{J\"{a}ger et~al\mbox{.}}{2019}]%
        {Jaeger2019}
\bibfield{author}{\bibinfo{person}{Lena~A. J\"{a}ger}, \bibinfo{person}{Silvia
  Makowski}, \bibinfo{person}{Paul Prasse}, \bibinfo{person}{Sascha Liehr},
  \bibinfo{person}{Maximilian Seidler}, {and} \bibinfo{person}{Tobias
  Scheffer}.} \bibinfo{year}{2019}\natexlab{}.
\newblock \showarticletitle{Deep Eyedentification: Biometric Identification
  Using Micro-Movements of the Eye}. In \bibinfo{booktitle}{\emph{Machine
  Learning and Knowledge Discovery in Databases: European Conference, ECML PKDD
  2019, W\"{u}rzburg, Germany, September 16–20, 2019, Proceedings, Part II}}
  (W\"{u}rzburg, Germany). \bibinfo{publisher}{Springer-Verlag},
  \bibinfo{address}{Berlin, Heidelberg}, \bibinfo{pages}{299–314}.
\newblock
\showISBNx{978-3-030-46146-1}
\urldef\tempurl%
\url{https://doi.org/10.1007/978-3-030-46147-8_18}
\showDOI{\tempurl}


\bibitem[\protect\citeauthoryear{Jiang and Zhao}{Jiang and Zhao}{2017}]%
        {jiang2017learning}
\bibfield{author}{\bibinfo{person}{Ming Jiang} {and} \bibinfo{person}{Qi
  Zhao}.} \bibinfo{year}{2017}\natexlab{}.
\newblock \showarticletitle{Learning visual attention to identify people with
  autism spectrum disorder}. In \bibinfo{booktitle}{\emph{Proceedings of IEEE
  International Conference on Computer Vision (ICCV)}}.
  \bibinfo{pages}{3267--3276}.
\newblock


\bibitem[\protect\citeauthoryear{Jäger, Kern, and Haller}{Jäger
  et~al\mbox{.}}{2021}]%
        {Jaeger2021b}
\bibfield{author}{\bibinfo{person}{Lena~A. Jäger}, \bibinfo{person}{Thomas
  Kern}, {and} \bibinfo{person}{Patrick Haller}.}
  \bibinfo{year}{2021}\natexlab{}.
\newblock \bibinfo{title}{Potsdam Textbook Corpus ({PoTeC}): Eye tracking data
  from experts and non-experts reading scientific texts}.
\newblock \bibinfo{howpublished}{DOI: 10.17605/OSF.IO/DN5HP}.
\newblock


\bibitem[\protect\citeauthoryear{Key, Venker, and Sandbank}{Key
  et~al\mbox{.}}{2020}]%
        {Key2020}
\bibfield{author}{\bibinfo{person}{Alexandra~P. Key},
  \bibinfo{person}{Courtney~E. Venker}, {and} \bibinfo{person}{Micheal~P.
  Sandbank}.} \bibinfo{year}{2020}\natexlab{}.
\newblock \showarticletitle{Psychophysiological and Eye-Tracking Markers of
  Speech and Language Processing in Neurodevelopmental Disorders: New Options
  for Difficult-to-Test Populations}.
\newblock \bibinfo{journal}{\emph{American Journal on Intellectual and
  Developmental Disabilities}} \bibinfo{volume}{125}, \bibinfo{number}{6}
  (\bibinfo{date}{nov} \bibinfo{year}{2020}), \bibinfo{pages}{465--474}.
\newblock
\urldef\tempurl%
\url{https://doi.org/10.1352/1944-7558-125.6.465}
\showDOI{\tempurl}


\bibitem[\protect\citeauthoryear{Kim, Wattenberg, Gilmer, Cai, Wexler, Viegas,
  and Sayres}{Kim et~al\mbox{.}}{2017}]%
        {Kim2017}
\bibfield{author}{\bibinfo{person}{Been Kim}, \bibinfo{person}{Martin
  Wattenberg}, \bibinfo{person}{Justin Gilmer}, \bibinfo{person}{Carrie Cai},
  \bibinfo{person}{James Wexler}, \bibinfo{person}{Fernanda Viegas}, {and}
  \bibinfo{person}{Rory Sayres}.} \bibinfo{year}{2017}\natexlab{}.
\newblock \showarticletitle{Interpretability Beyond Feature Attribution:
  Quantitative Testing with Concept Activation Vectors (TCAV)}.
\newblock  (\bibinfo{year}{2017}).
\newblock
\urldef\tempurl%
\url{https://doi.org/10.48550/ARXIV.1711.11279}
\showDOI{\tempurl}


\bibitem[\protect\citeauthoryear{Kohlbrenner, Bauer, Nakajima, Binder, Samek,
  and Lapuschkin}{Kohlbrenner et~al\mbox{.}}{2020}]%
        {Kohlbrenner2020}
\bibfield{author}{\bibinfo{person}{Maximilian Kohlbrenner},
  \bibinfo{person}{Alexander Bauer}, \bibinfo{person}{Shinichi Nakajima},
  \bibinfo{person}{Alexander Binder}, \bibinfo{person}{Wojciech Samek}, {and}
  \bibinfo{person}{Sebastian Lapuschkin}.} \bibinfo{year}{2020}\natexlab{}.
\newblock \showarticletitle{Towards Best Practice in Explaining Neural Network
  Decisions with LRP}.
\newblock \bibinfo{journal}{\emph{International Joint Conference on Neural
  Networks}} (\bibinfo{year}{2020}), \bibinfo{pages}{1--7}.
\newblock


\bibitem[\protect\citeauthoryear{Kokhlikyan, Miglani, Martin, Wang, Alsallakh,
  Reynolds, Melnikov, Kliushkina, Araya, Yan, and
  Reblitz-Richardson}{Kokhlikyan et~al\mbox{.}}{2020}]%
        {Kokhlikyan2020}
\bibfield{author}{\bibinfo{person}{Narine Kokhlikyan}, \bibinfo{person}{Vivek
  Miglani}, \bibinfo{person}{Miguel Martin}, \bibinfo{person}{Edward Wang},
  \bibinfo{person}{Bilal Alsallakh}, \bibinfo{person}{Jonathan Reynolds},
  \bibinfo{person}{Alexander Melnikov}, \bibinfo{person}{Natalia Kliushkina},
  \bibinfo{person}{Carlos Araya}, \bibinfo{person}{Siqi Yan}, {and}
  \bibinfo{person}{Orion Reblitz-Richardson}.} \bibinfo{year}{2020}\natexlab{}.
\newblock \showarticletitle{Captum: A unified and generic model
  interpretability library for {PyTorch}}.
\newblock \bibinfo{journal}{\emph{arXiv preprint arXiv:2009.07896}}
  (\bibinfo{year}{2020}).
\newblock


\bibitem[\protect\citeauthoryear{Krakowczyk, Reich, Prasse, Lapuschkin,
  J{\"a}ger, and Scheffer}{Krakowczyk et~al\mbox{.}}{2022}]%
        {Krakowczyk2022}
\bibfield{author}{\bibinfo{person}{Daniel Krakowczyk},
  \bibinfo{person}{David~Robert Reich}, \bibinfo{person}{Paul Prasse},
  \bibinfo{person}{Sebastian Lapuschkin}, \bibinfo{person}{Lena~Ann J{\"a}ger},
  {and} \bibinfo{person}{Tobias Scheffer}.} \bibinfo{year}{2022}\natexlab{}.
\newblock \showarticletitle{Selection of {XAI} Methods Matters: Evaluation of
  Feature Attribution Methods for Oculomotoric Biometric Identification}. In
  \bibinfo{booktitle}{\emph{NeuRIPS 2022 Workshop on Gaze Meets ML}}.
\newblock
\urldef\tempurl%
\url{https://openreview.net/forum?id=GOLdDAP2AtI}
\showURL{%
\tempurl}


\bibitem[\protect\citeauthoryear{Krakowczyk, Reich, Chwastek, Jakobi, Prasse,
  Süss, Turuta, Kasprowski, and Jäger}{Krakowczyk et~al\mbox{.}}{2023}]%
        {pymovements}
\bibfield{author}{\bibinfo{person}{Daniel~G. Krakowczyk},
  \bibinfo{person}{David~R. Reich}, \bibinfo{person}{Jakob Chwastek},
  \bibinfo{person}{Deborah~N. Jakobi}, \bibinfo{person}{Paul Prasse},
  \bibinfo{person}{Assunta Süss}, \bibinfo{person}{Oleksii Turuta},
  \bibinfo{person}{Paweł Kasprowski}, {and} \bibinfo{person}{Lena~A. Jäger}.}
  \bibinfo{year}{2023}\natexlab{}.
\newblock \showarticletitle{pymovements: A Python Package for Processing Eye
  Movement Data}. In \bibinfo{booktitle}{\emph{2023 Symposium on Eye Tracking
  Research and Applications}} (Tubingen, Germany) \emph{(\bibinfo{series}{ETRA
  '23})}. \bibinfo{publisher}{Association for Computing Machinery},
  \bibinfo{address}{New York, NY, USA}.
\newblock
\showISBNx{979-8-4007-0150-4/23/05}
\urldef\tempurl%
\url{https://doi.org/10.1145/3588015.3590134}
\showDOI{\tempurl}


\bibitem[\protect\citeauthoryear{Krizhevsky, Sutskever, and Hinton}{Krizhevsky
  et~al\mbox{.}}{2012}]%
        {Krizhevsky2012}
\bibfield{author}{\bibinfo{person}{Alex Krizhevsky}, \bibinfo{person}{Ilya
  Sutskever}, {and} \bibinfo{person}{Geoffrey~E Hinton}.}
  \bibinfo{year}{2012}\natexlab{}.
\newblock \showarticletitle{ImageNet Classification with Deep Convolutional
  Neural Networks}. In \bibinfo{booktitle}{\emph{Advances in Neural Information
  Processing Systems}}, \bibfield{editor}{\bibinfo{person}{F.~Pereira},
  \bibinfo{person}{C.J. Burges}, \bibinfo{person}{L.~Bottou}, {and}
  \bibinfo{person}{K.Q. Weinberger}} (Eds.), Vol.~\bibinfo{volume}{25}.
  \bibinfo{publisher}{Curran Associates, Inc.}
\newblock
\urldef\tempurl%
\url{https://proceedings.neurips.cc/paper/2012/file/c399862d3b9d6b76c8436e924a68c45b-Paper.pdf}
\showURL{%
\tempurl}


\bibitem[\protect\citeauthoryear{Kumar, Howlader, Garcia, Weiskopf, and
  Mueller}{Kumar et~al\mbox{.}}{2020}]%
        {Kumar2020}
\bibfield{author}{\bibinfo{person}{Ayush Kumar}, \bibinfo{person}{Prantik
  Howlader}, \bibinfo{person}{Rafael Garcia}, \bibinfo{person}{Daniel
  Weiskopf}, {and} \bibinfo{person}{Klaus Mueller}.}
  \bibinfo{year}{2020}\natexlab{}.
\newblock \showarticletitle{Challenges in Interpretability of Neural Networks
  for Eye Movement Data}. In \bibinfo{booktitle}{\emph{ACM Symposium on Eye
  Tracking Research and Applications}} (Stuttgart, Germany)
  \emph{(\bibinfo{series}{ETRA '20 Short Papers})}.
  \bibinfo{publisher}{Association for Computing Machinery},
  \bibinfo{address}{New York, NY, USA}, Article \bibinfo{articleno}{12},
  \bibinfo{numpages}{5}~pages.
\newblock
\showISBNx{9781450371346}
\urldef\tempurl%
\url{https://doi.org/10.1145/3379156.3391361}
\showDOI{\tempurl}


\bibitem[\protect\citeauthoryear{Lohr, Griffith, and Komogortsev}{Lohr
  et~al\mbox{.}}{2021}]%
        {Lohr2021}
\bibfield{author}{\bibinfo{person}{Dillon Lohr}, \bibinfo{person}{Henry
  Griffith}, {and} \bibinfo{person}{Oleg~V Komogortsev}.}
  \bibinfo{year}{2021}\natexlab{}.
\newblock \bibinfo{title}{Eye Know You: Metric Learning for End-to-end
  Biometric Authentication Using Eye Movements from a Longitudinal Dataset}.
\newblock
\newblock
\urldef\tempurl%
\url{https://doi.org/10.48550/ARXIV.2104.10489}
\showDOI{\tempurl}


\bibitem[\protect\citeauthoryear{Lohr and Komogortsev}{Lohr and
  Komogortsev}{2022}]%
        {Lohr2022}
\bibfield{author}{\bibinfo{person}{Dillon Lohr} {and} \bibinfo{person}{Oleg~V
  Komogortsev}.} \bibinfo{year}{2022}\natexlab{}.
\newblock \showarticletitle{{Eye Know You Too}: Toward Viable End-to-End Eye
  Movement Biometrics for User Authentication}.
\newblock \bibinfo{journal}{\emph{IEEE Transactions on Information Forensics
  and Security}}  \bibinfo{volume}{17} (\bibinfo{year}{2022}),
  \bibinfo{pages}{3151--3164}.
\newblock


\bibitem[\protect\citeauthoryear{Makowski, Jäger, Prasse, and
  Scheffer}{Makowski et~al\mbox{.}}{2020}]%
        {Makowski2020b}
\bibfield{author}{\bibinfo{person}{Silvia Makowski}, \bibinfo{person}{Lena~A.
  Jäger}, \bibinfo{person}{Paul Prasse}, {and} \bibinfo{person}{Tobias
  Scheffer}.} \bibinfo{year}{2020}\natexlab{}.
\newblock \bibinfo{title}{{JuDo1000} Eye Tracking Data Set}.
\newblock \bibinfo{howpublished}{DOI: 10.17605/OSF.IO/5ZPVK}. ,
  \bibinfo{numpages}{10}~pages.
\newblock


\bibitem[\protect\citeauthoryear{Makowski, Prasse, Reich, Krakowczyk, Jäger,
  and Scheffer}{Makowski et~al\mbox{.}}{2021}]%
        {Makowski2021}
\bibfield{author}{\bibinfo{person}{Silvia Makowski}, \bibinfo{person}{Paul
  Prasse}, \bibinfo{person}{David~R. Reich}, \bibinfo{person}{Daniel
  Krakowczyk}, \bibinfo{person}{Lena~A. Jäger}, {and} \bibinfo{person}{Tobias
  Scheffer}.} \bibinfo{year}{2021}\natexlab{}.
\newblock \showarticletitle{{DeepEyedentificationLive}: Oculomotoric Biometric
  Identification and Presentation-Attack Detection Using Deep Neural Networks}.
\newblock \bibinfo{journal}{\emph{IEEE Transactions on Biometrics, Behavior,
  and Identity Science}} \bibinfo{volume}{3}, \bibinfo{number}{4}
  (\bibinfo{year}{2021}), \bibinfo{pages}{506--518}.
\newblock


\bibitem[\protect\citeauthoryear{Martinez-Conde, Macknik, and
  Hubel}{Martinez-Conde et~al\mbox{.}}{2004}]%
        {Martinez-Conde2004}
\bibfield{author}{\bibinfo{person}{Susana Martinez-Conde},
  \bibinfo{person}{Stephen~L. Macknik}, {and} \bibinfo{person}{David~H.
  Hubel}.} \bibinfo{year}{2004}\natexlab{}.
\newblock \showarticletitle{The role of fixational eye movements in visual
  perception}.
\newblock \bibinfo{journal}{\emph{Nature Reviews Neuroscience}}
  \bibinfo{volume}{5} (\bibinfo{year}{2004}), \bibinfo{pages}{229–240}.
\newblock


\bibitem[\protect\citeauthoryear{Martinez-Conde, Macknik, Troncoso, and
  Dyar}{Martinez-Conde et~al\mbox{.}}{2006}]%
        {Martinez-Conde2006}
\bibfield{author}{\bibinfo{person}{Susana Martinez-Conde},
  \bibinfo{person}{Stephen~L. Macknik}, \bibinfo{person}{Xoana~G. Troncoso},
  {and} \bibinfo{person}{Thomas~A. Dyar}.} \bibinfo{year}{2006}\natexlab{}.
\newblock \showarticletitle{Microsaccades Counteract Visual Fading during
  Fixation}.
\newblock \bibinfo{journal}{\emph{Neuron}}  \bibinfo{volume}{49}
  (\bibinfo{year}{2006}), \bibinfo{pages}{297--305}.
\newblock


\bibitem[\protect\citeauthoryear{Molnar}{Molnar}{2022}]%
        {Molnar2022}
\bibfield{author}{\bibinfo{person}{Christoph Molnar}.}
  \bibinfo{year}{2022}\natexlab{}.
\newblock \bibinfo{booktitle}{\emph{Interpretable Machine Learning}
  (\bibinfo{edition}{2} ed.)}.
\newblock
\urldef\tempurl%
\url{https://christophm.github.io/interpretable-ml-book}
\showURL{%
\tempurl}


\bibitem[\protect\citeauthoryear{Montavon, Binder, Lapuschkin, Samek, and
  M{\"u}ller}{Montavon et~al\mbox{.}}{2019}]%
        {Montavon2019}
\bibfield{author}{\bibinfo{person}{Gr{\'e}goire Montavon},
  \bibinfo{person}{Alexander Binder}, \bibinfo{person}{Sebastian Lapuschkin},
  \bibinfo{person}{Wojciech Samek}, {and} \bibinfo{person}{Klaus-Robert
  M{\"u}ller}.} \bibinfo{year}{2019}\natexlab{}.
\newblock \bibinfo{booktitle}{\emph{Layer-Wise Relevance Propagation: An
  Overview}}.
\newblock \bibinfo{publisher}{Springer International Publishing},
  \bibinfo{pages}{193--209}.
\newblock


\bibitem[\protect\citeauthoryear{Montavon, Lapuschkin, Binder, Samek, and
  Müller}{Montavon et~al\mbox{.}}{2017}]%
        {Montavon2017}
\bibfield{author}{\bibinfo{person}{Grégoire Montavon},
  \bibinfo{person}{Sebastian Lapuschkin}, \bibinfo{person}{Alexander Binder},
  \bibinfo{person}{Wojciech Samek}, {and} \bibinfo{person}{Klaus-Robert
  Müller}.} \bibinfo{year}{2017}\natexlab{}.
\newblock \showarticletitle{Explaining nonlinear classification decisions with
  deep Taylor decomposition}.
\newblock \bibinfo{journal}{\emph{Pattern Recognition}}  \bibinfo{volume}{65}
  (\bibinfo{year}{2017}), \bibinfo{pages}{211--222}.
\newblock


\bibitem[\protect\citeauthoryear{Montavon, Samek, and Müller}{Montavon
  et~al\mbox{.}}{2018}]%
        {Montavon2018}
\bibfield{author}{\bibinfo{person}{Grégoire Montavon},
  \bibinfo{person}{Wojciech Samek}, {and} \bibinfo{person}{Klaus-Robert
  Müller}.} \bibinfo{year}{2018}\natexlab{}.
\newblock \showarticletitle{Methods for interpreting and understanding deep
  neural networks}.
\newblock \bibinfo{journal}{\emph{Digital Signal Processing}}
  \bibinfo{volume}{73} (\bibinfo{year}{2018}), \bibinfo{pages}{1--15}.
\newblock


\bibitem[\protect\citeauthoryear{Nassif, Shahin, Attili, Azzeh, and
  Shaalan}{Nassif et~al\mbox{.}}{2019}]%
        {Nassif2019}
\bibfield{author}{\bibinfo{person}{Ali~Bou Nassif}, \bibinfo{person}{Ismail
  Shahin}, \bibinfo{person}{Imtinan Attili}, \bibinfo{person}{Mohammad Azzeh},
  {and} \bibinfo{person}{Khaled Shaalan}.} \bibinfo{year}{2019}\natexlab{}.
\newblock \showarticletitle{Speech Recognition Using Deep Neural Networks: A
  Systematic Review}.
\newblock \bibinfo{journal}{\emph{IEEE Access}}  \bibinfo{volume}{7}
  (\bibinfo{year}{2019}), \bibinfo{pages}{19143--19165}.
\newblock
\urldef\tempurl%
\url{https://doi.org/10.1109/ACCESS.2019.2896880}
\showDOI{\tempurl}


\bibitem[\protect\citeauthoryear{Nyström and Holmqvist}{Nyström and
  Holmqvist}{2010}]%
        {NystromHolmqvist2010}
\bibfield{author}{\bibinfo{person}{Marcus Nyström} {and}
  \bibinfo{person}{Kenneth Holmqvist}.} \bibinfo{year}{2010}\natexlab{}.
\newblock \showarticletitle{An adaptive algorithm for fixation, saccade, and
  glissade detection in eyetracking data.}
\newblock \bibinfo{journal}{\emph{Behavior Research Methods}}
  \bibinfo{volume}{42}, \bibinfo{number}{1} (\bibinfo{year}{2010}),
  \bibinfo{pages}{188--204}.
\newblock


\bibitem[\protect\citeauthoryear{Pedregosa, Varoquaux, Gramfort, Michel,
  Thirion, Grisel, Blondel, Prettenhofer, Weiss, Dubourg, Vanderplas, Passos,
  Cournapeau, Brucher, Perrot, and Duchesnay}{Pedregosa et~al\mbox{.}}{2011}]%
        {scikit-learn}
\bibfield{author}{\bibinfo{person}{F. Pedregosa}, \bibinfo{person}{G.
  Varoquaux}, \bibinfo{person}{A. Gramfort}, \bibinfo{person}{V. Michel},
  \bibinfo{person}{B. Thirion}, \bibinfo{person}{O. Grisel},
  \bibinfo{person}{M. Blondel}, \bibinfo{person}{P. Prettenhofer},
  \bibinfo{person}{R. Weiss}, \bibinfo{person}{V. Dubourg}, \bibinfo{person}{J.
  Vanderplas}, \bibinfo{person}{A. Passos}, \bibinfo{person}{D. Cournapeau},
  \bibinfo{person}{M. Brucher}, \bibinfo{person}{M. Perrot}, {and}
  \bibinfo{person}{E. Duchesnay}.} \bibinfo{year}{2011}\natexlab{}.
\newblock \showarticletitle{Scikit-learn: Machine learning in {P}ython}.
\newblock \bibinfo{journal}{\emph{Journal of Machine Learning Research}}
  \bibinfo{volume}{12} (\bibinfo{year}{2011}), \bibinfo{pages}{2825--2830}.
\newblock


\bibitem[\protect\citeauthoryear{Raatikainen, Hautala, Loberg,
  K{\"a}rkk{\"a}inen, Lepp{\"a}nen, and Nieminen}{Raatikainen
  et~al\mbox{.}}{2021}]%
        {raatikainen2021detection}
\bibfield{author}{\bibinfo{person}{Peter Raatikainen}, \bibinfo{person}{Jarkko
  Hautala}, \bibinfo{person}{Otto Loberg}, \bibinfo{person}{Tommi
  K{\"a}rkk{\"a}inen}, \bibinfo{person}{Paavo Lepp{\"a}nen}, {and}
  \bibinfo{person}{Paavo Nieminen}.} \bibinfo{year}{2021}\natexlab{}.
\newblock \showarticletitle{Detection of developmental dyslexia with machine
  learning using eye movement data}.
\newblock \bibinfo{journal}{\emph{Array}}  \bibinfo{volume}{12}
  (\bibinfo{year}{2021}), \bibinfo{pages}{100087}.
\newblock


\bibitem[\protect\citeauthoryear{Rayner, Ashby, Pollatsek, and Reichle}{Rayner
  et~al\mbox{.}}{2004}]%
        {Rayner2004}
\bibfield{author}{\bibinfo{person}{Keith Rayner}, \bibinfo{person}{Jane Ashby},
  \bibinfo{person}{Alexander Pollatsek}, {and} \bibinfo{person}{Erik~D.
  Reichle}.} \bibinfo{year}{2004}\natexlab{}.
\newblock \showarticletitle{The Effects of Frequency and Predictability on Eye
  Fixations in Reading: Implications for the E-Z Reader Model.}
\newblock \bibinfo{journal}{\emph{Journal of Experimental Psychology: Human
  Perception and Performance}} \bibinfo{volume}{30}, \bibinfo{number}{4}
  (\bibinfo{year}{2004}), \bibinfo{pages}{720--732}.
\newblock
\urldef\tempurl%
\url{https://doi.org/10.1037/0096-1523.30.4.720}
\showDOI{\tempurl}


\bibitem[\protect\citeauthoryear{Rayner and Pollatsek}{Rayner and
  Pollatsek}{1983}]%
        {Rayner1983}
\bibfield{author}{\bibinfo{person}{Keith Rayner} {and}
  \bibinfo{person}{Alexander Pollatsek}.} \bibinfo{year}{1983}\natexlab{}.
\newblock \showarticletitle{Is visual information integrated across saccades?}
\newblock \bibinfo{journal}{\emph{Perception \& Psychophysics}}
  \bibinfo{volume}{34}, \bibinfo{number}{1} (\bibinfo{date}{jan}
  \bibinfo{year}{1983}), \bibinfo{pages}{39--48}.
\newblock
\urldef\tempurl%
\url{https://doi.org/10.3758/bf03205894}
\showDOI{\tempurl}


\bibitem[\protect\citeauthoryear{Rigas, Friedman, and Komogortsev}{Rigas
  et~al\mbox{.}}{2018}]%
        {rigas2018study}
\bibfield{author}{\bibinfo{person}{Ioannis Rigas}, \bibinfo{person}{Lee
  Friedman}, {and} \bibinfo{person}{Oleg Komogortsev}.}
  \bibinfo{year}{2018}\natexlab{}.
\newblock \showarticletitle{Study of an extensive set of eye movement features:
  Extraction methods and statistical analysis}.
\newblock \bibinfo{journal}{\emph{Journal of Eye Movement Research}}
  \bibinfo{volume}{11}, \bibinfo{number}{1} (\bibinfo{year}{2018}).
\newblock


\bibitem[\protect\citeauthoryear{Rigas, Komogortsev, and Shadmehr}{Rigas
  et~al\mbox{.}}{2016}]%
        {rigas2016biometric}
\bibfield{author}{\bibinfo{person}{Ioannis Rigas}, \bibinfo{person}{Oleg
  Komogortsev}, {and} \bibinfo{person}{Reza Shadmehr}.}
  \bibinfo{year}{2016}\natexlab{}.
\newblock \showarticletitle{Biometric recognition via eye movements: Saccadic
  vigor and acceleration cues}.
\newblock \bibinfo{journal}{\emph{ACM Transactions on Applied Perception}}
  \bibinfo{volume}{13}, \bibinfo{number}{2} (\bibinfo{year}{2016}),
  \bibinfo{pages}{1--21}.
\newblock


\bibitem[\protect\citeauthoryear{Salvucci and Goldberg}{Salvucci and
  Goldberg}{2000}]%
        {Salvucci2000}
\bibfield{author}{\bibinfo{person}{Dario~D. Salvucci} {and}
  \bibinfo{person}{Joseph~H. Goldberg}.} \bibinfo{year}{2000}\natexlab{}.
\newblock \showarticletitle{Identifying Fixations and Saccades in Eye-Tracking
  Protocols}. In \bibinfo{booktitle}{\emph{ETRA '00}} (Palm Beach Gardens,
  Florida, USA) \emph{(\bibinfo{series}{ETRA '00})}.
  \bibinfo{publisher}{Association for Computing Machinery},
  \bibinfo{address}{New York, NY, USA}, \bibinfo{pages}{71–78}.
\newblock
\showISBNx{1581132808}
\urldef\tempurl%
\url{https://doi.org/10.1145/355017.355028}
\showDOI{\tempurl}


\bibitem[\protect\citeauthoryear{Savitzky and Golay}{Savitzky and
  Golay}{1964}]%
        {SavitzkyGolay1964}
\bibfield{author}{\bibinfo{person}{Abraham. Savitzky} {and}
  \bibinfo{person}{M.~J.~E. Golay}.} \bibinfo{year}{1964}\natexlab{}.
\newblock \showarticletitle{Smoothing and Differentiation of Data by Simplified
  Least Squares Procedures.}
\newblock \bibinfo{journal}{\emph{Analytical Chemistry}} \bibinfo{volume}{36},
  \bibinfo{number}{8} (\bibinfo{year}{1964}), \bibinfo{pages}{1627--1639}.
\newblock


\bibitem[\protect\citeauthoryear{Shrikumar, Greenside, and Kundaje}{Shrikumar
  et~al\mbox{.}}{2017}]%
        {Shrikumar2017}
\bibfield{author}{\bibinfo{person}{Avanti Shrikumar}, \bibinfo{person}{Peyton
  Greenside}, {and} \bibinfo{person}{Anshul Kundaje}.}
  \bibinfo{year}{2017}\natexlab{}.
\newblock \showarticletitle{Learning Important Features Through Propagating
  Activation Differences}. In \bibinfo{booktitle}{\emph{Proceedings of the 34th
  International Conference on Machine Learning (ICML)}}
  \emph{(\bibinfo{series}{Proceedings of Machine Learning Research},
  Vol.~\bibinfo{volume}{70})}, \bibfield{editor}{\bibinfo{person}{Doina Precup}
  {and} \bibinfo{person}{Yee~Whye Teh}} (Eds.). \bibinfo{publisher}{PMLR},
  \bibinfo{pages}{3145--3153}.
\newblock


\bibitem[\protect\citeauthoryear{Shrikumar, Greenside, Shcherbina, and
  Kundaje}{Shrikumar et~al\mbox{.}}{2016}]%
        {Shrikumar2016}
\bibfield{author}{\bibinfo{person}{Avanti Shrikumar}, \bibinfo{person}{Peyton
  Greenside}, \bibinfo{person}{Anna Shcherbina}, {and} \bibinfo{person}{Anshul
  Kundaje}.} \bibinfo{year}{2016}\natexlab{}.
\newblock \showarticletitle{Not Just a Black Box: Learning Important Features
  Through Propagating Activation Differences}.
\newblock \bibinfo{journal}{\emph{arXiv preprint arXiv:1605.01713}}
  (\bibinfo{year}{2016}).
\newblock


\bibitem[\protect\citeauthoryear{Startsev and Zemblys}{Startsev and
  Zemblys}{2022}]%
        {Startsev2022}
\bibfield{author}{\bibinfo{person}{Mikhail Startsev} {and}
  \bibinfo{person}{Raimondas Zemblys}.} \bibinfo{year}{2022}\natexlab{}.
\newblock \showarticletitle{Evaluating Eye Movement Event Detection: A Review
  of the State of the Art}.
\newblock \bibinfo{journal}{\emph{Behavior Research Methods}}
  (\bibinfo{date}{jun} \bibinfo{year}{2022}).
\newblock
\urldef\tempurl%
\url{https://doi.org/10.3758/s13428-021-01763-7}
\showDOI{\tempurl}


\bibitem[\protect\citeauthoryear{Sturmfels, Lundberg, and Lee}{Sturmfels
  et~al\mbox{.}}{2020}]%
        {Sturmfels2020}
\bibfield{author}{\bibinfo{person}{Pascal Sturmfels}, \bibinfo{person}{Scott
  Lundberg}, {and} \bibinfo{person}{Su-In Lee}.}
  \bibinfo{year}{2020}\natexlab{}.
\newblock \showarticletitle{Visualizing the Impact of Feature Attribution
  Baselines}.
\newblock \bibinfo{journal}{\emph{Distill}} \bibinfo{volume}{5},
  \bibinfo{number}{1} (\bibinfo{year}{2020}).
\newblock


\bibitem[\protect\citeauthoryear{Sundararajan, Taly, and Yan}{Sundararajan
  et~al\mbox{.}}{2017}]%
        {Sundararajan2017}
\bibfield{author}{\bibinfo{person}{Mukund Sundararajan}, \bibinfo{person}{Ankur
  Taly}, {and} \bibinfo{person}{Qiqi Yan}.} \bibinfo{year}{2017}\natexlab{}.
\newblock \showarticletitle{Axiomatic Attribution for Deep Networks}. In
  \bibinfo{booktitle}{\emph{Proceedings of the 34th International Conference on
  Machine Learning (ICML)}} \emph{(\bibinfo{series}{Proceedings of Machine
  Learning Research}, Vol.~\bibinfo{volume}{70})},
  \bibfield{editor}{\bibinfo{person}{Doina Precup} {and}
  \bibinfo{person}{Yee~Whye Teh}} (Eds.). \bibinfo{publisher}{PMLR},
  \bibinfo{pages}{3319--3328}.
\newblock


\bibitem[\protect\citeauthoryear{Theiner, M\"uller-Budack, and Ewerth}{Theiner
  et~al\mbox{.}}{2022}]%
        {Theiner2022}
\bibfield{author}{\bibinfo{person}{Jonas Theiner}, \bibinfo{person}{Eric
  M\"uller-Budack}, {and} \bibinfo{person}{Ralph Ewerth}.}
  \bibinfo{year}{2022}\natexlab{}.
\newblock \showarticletitle{Interpretable Semantic Photo Geolocation}. In
  \bibinfo{booktitle}{\emph{Proceedings of the IEEE/CVF Winter Conference on
  Applications of Computer Vision (WACV)}}. \bibinfo{pages}{750--760}.
\newblock


\bibitem[\protect\citeauthoryear{Voulodimos, Doulamis, Doulamis,
  Protopapadakis, and Andina}{Voulodimos et~al\mbox{.}}{2018}]%
        {Voulodimos2018}
\bibfield{author}{\bibinfo{person}{Athanasios Voulodimos},
  \bibinfo{person}{Nikolaos Doulamis}, \bibinfo{person}{Anastasios Doulamis},
  \bibinfo{person}{Eftychios Protopapadakis}, {and} \bibinfo{person}{Diego
  Andina}.} \bibinfo{year}{2018}\natexlab{}.
\newblock \showarticletitle{Deep Learning for Computer Vision: A Brief Review}.
\newblock \bibinfo{journal}{\emph{Intell. Neuroscience}}
  \bibinfo{volume}{2018} (\bibinfo{date}{jan} \bibinfo{year}{2018}), 13.
\newblock
\showISSN{1687-5265}
\urldef\tempurl%
\url{https://doi.org/10.1155/2018/7068349}
\showDOI{\tempurl}


\bibitem[\protect\citeauthoryear{Williford, May, and Byrne}{Williford
  et~al\mbox{.}}{2020}]%
        {Williford2020}
\bibfield{author}{\bibinfo{person}{Jonathan~R. Williford},
  \bibinfo{person}{Brandon~B. May}, {and} \bibinfo{person}{Jeffrey Byrne}.}
  \bibinfo{year}{2020}\natexlab{}.
\newblock \bibinfo{title}{Explainable Face Recognition}.
\newblock
\newblock
\urldef\tempurl%
\url{https://doi.org/10.48550/ARXIV.2008.00916}
\showDOI{\tempurl}


\end{thebibliography}

\begin{appendix}

\section{Dataset Statistics}
\label{apx:datasets}
Table~\ref{tab:datasets} provides a brief summary of dataset properties.

\begin{table}[h!]
\centering
\begin{tabular}{lrrl}
\toprule
  & \# of subjects & \# of sessions & stimulus \\[0.5ex]
 \hline
 & \\[-2ex]
 GazeBaze & 100 & 4 & random dots, text, video \\
 JuDo1000 & 150 & 4 & random dots \\
 PoTeC & 75 & 1 & text \\[1ex]
\bottomrule
\end{tabular}
\caption{Dataset properties.}
\label{tab:datasets}
\end{table}

\section{Event Detection}
\label{apx:event-detection-parameters}

Table~\ref{tab:event-detection-parameters} lists all parameters used in the event detection process.

\begin{table}[h!]
\centering
\begin{tabular}{lrl} 
 \toprule
 Parameter & Value \\[0.5ex]
 \hline
 & \\[-2ex]
 Max fix. velocity & \SI{20}{\degree/\second} &\cite{Salvucci2000} \\
 Min fix. duration & \SI{40}{\ms} &\cite{NystromHolmqvist2010} \\
 Max fix. dispersion & $2.7\mskip3mu\degree$ &\cite{Andersson2016}  \\[1ex]
 Sacc. threshold factor & 6 &\cite{EngbertKliegl2003} \\ 
 Min sacc. duration & \SI{9}{\ms} &\cite{Dombrow2018} \\ 
 Max sacc. duration & \SI{100}{\ms} &\cite{Dombrow2018} \\
 Min sacc. peak vel. & \SI{35}{\degree/\second} &\cite{NystromHolmqvist2010}  \\
 Max sacc. peak vel. & \SI{1000}{\degree/\second} &\cite{NystromHolmqvist2010}  \\[0.5ex]
\bottomrule
\end{tabular}
\caption{Parameter values for event detection algorithms.}
\label{tab:event-detection-parameters}
\end{table}

\section{Model Accuracy}
\label{apx:model-accuracy}

\begin{figure}[h!]
\centering
\includegraphics[width=\linewidth]{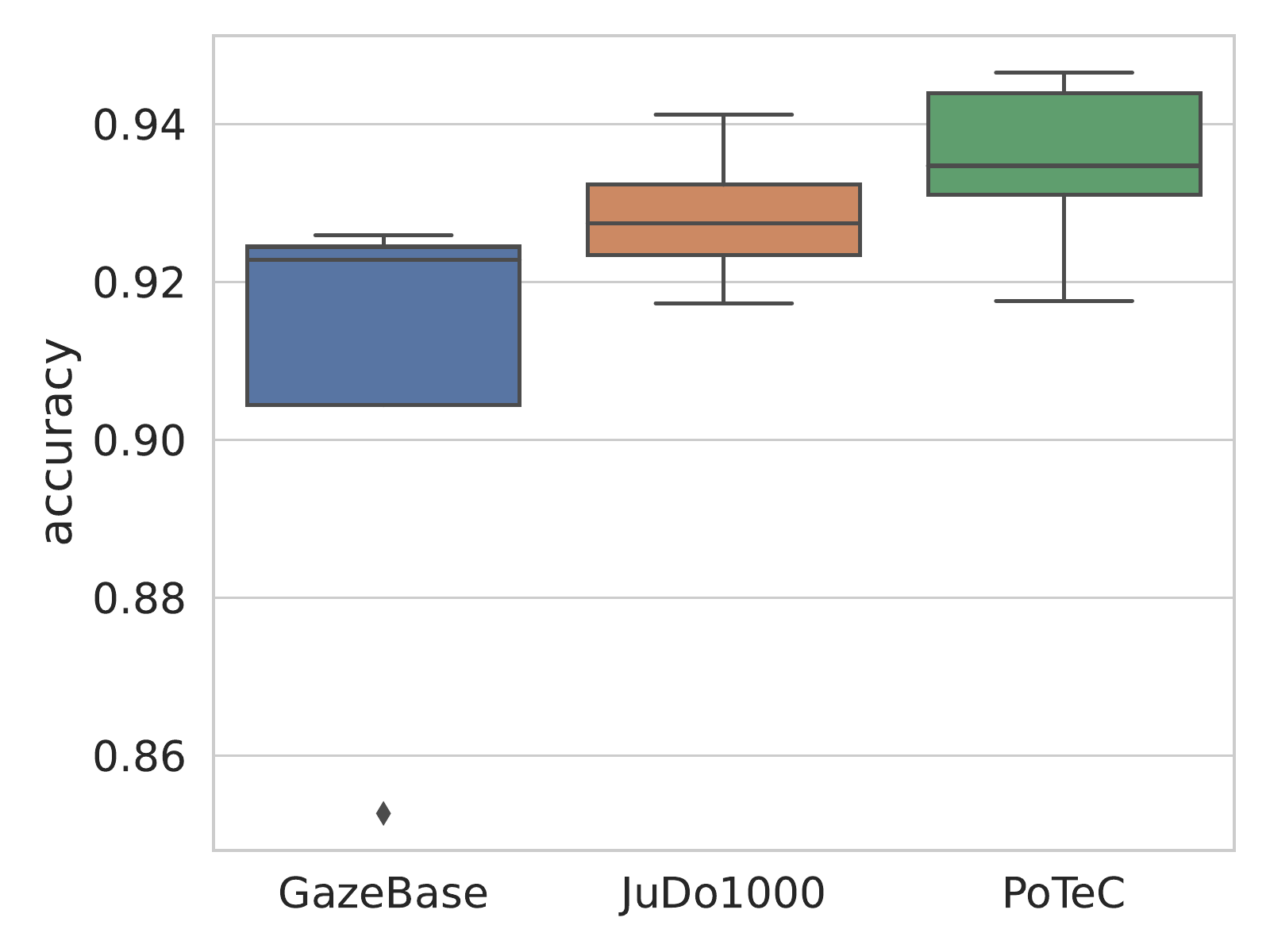}
\caption{Model accuracy for different datasets.}
\label{fig:model-accuracy}
\end{figure}

Figure~\ref{fig:model-accuracy} presents model accuracies across datasets.

\section{Segmentation Sizes}
This section contains the relative segmentation sizes for each of the experiments from the main paper.

\subsection{Saccades and Fixations}
\label{apx:event-type-sizes}

\begin{figure}[h!]
\centering
\includegraphics[width=\linewidth]{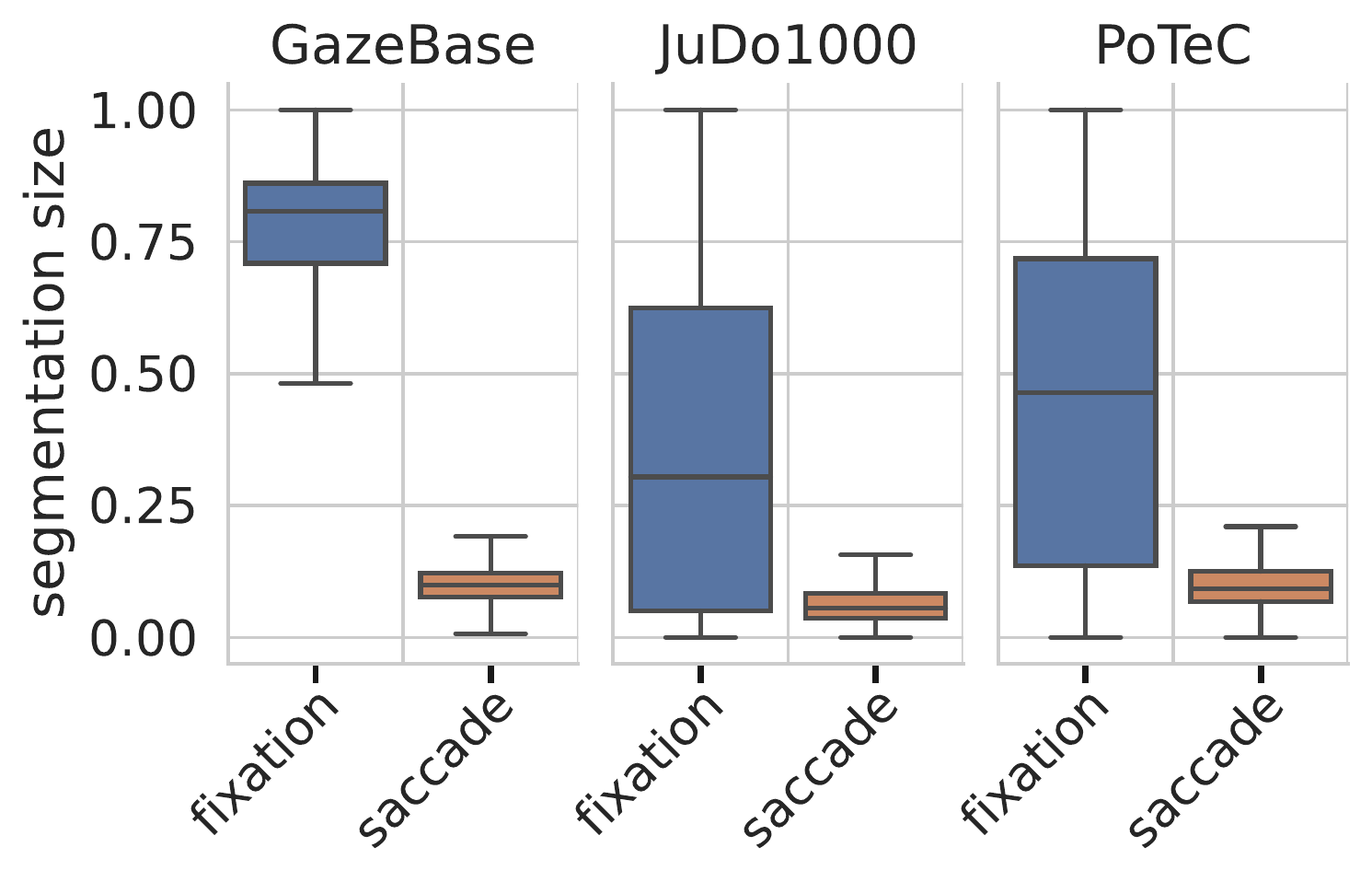}
\caption{Segmentation sizes for saccades and fixations.}
\label{fig:event-type-sizes}
\end{figure}

Figure~\ref{fig:event-type-sizes} contains the relative segmentation sizes of saccades and fixations.

\subsection{Saccade Sub-events}
\label{apx:event-dissection-sizes}

\begin{figure}[h!]
\centering
\includegraphics[width=\linewidth]{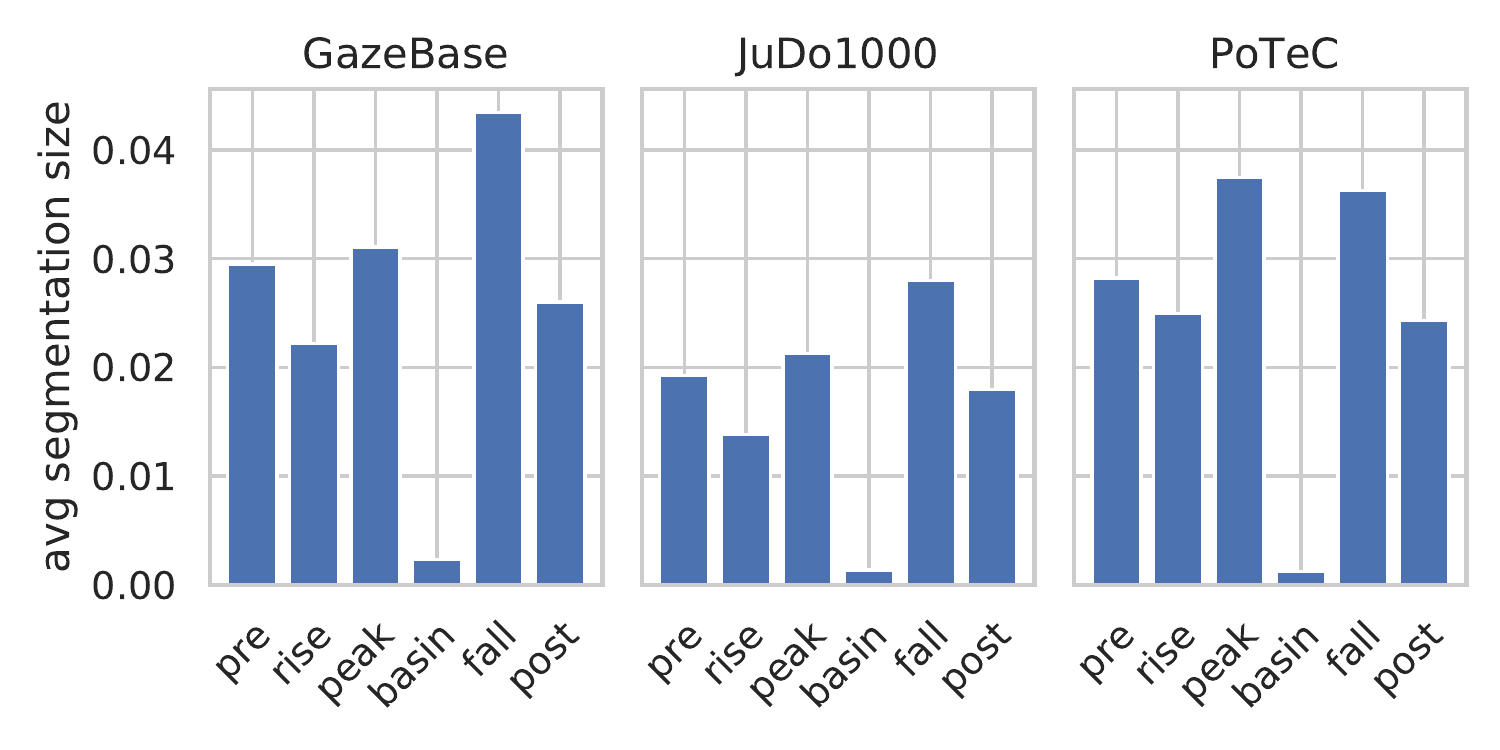}
\caption{Segmentation sizes for sub-events of saccades.}
\label{fig:event-dissection-sizes}
\end{figure}

Figure~\ref{fig:event-dissection-sizes} contains the relative segmentation sizes of saccade sub-events.

\subsection{Event Properties}
\label{apx:event-property-binning-saccade-amplitude}

Figure~\ref{fig:event-property-binning-saccade-amplitude} contains the relative segmentation sizes across different event properties.

\begin{figure*}
\centering
\subfigure[Property binning by saccade duration.]{\includegraphics[width=0.99\linewidth]{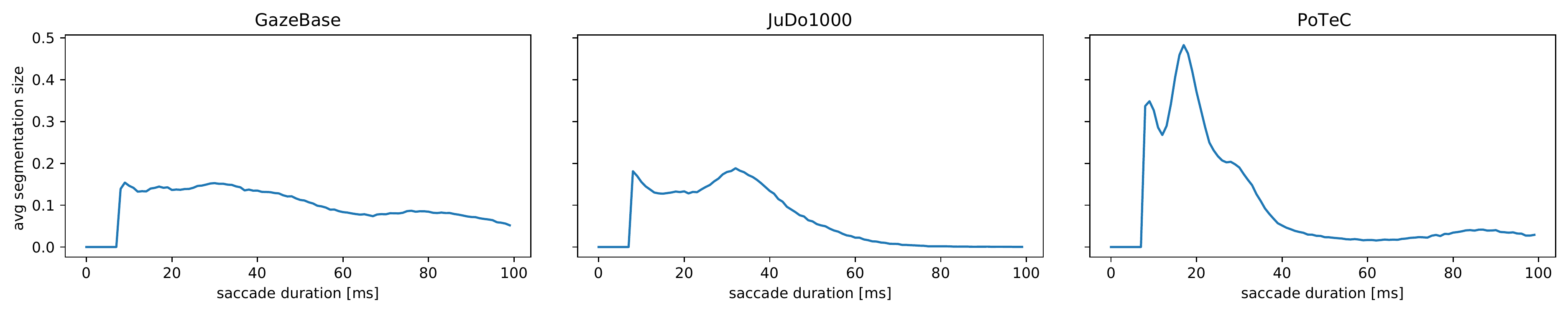}\label{fig:event-property-binning-saccade-duration-size}}\\
\subfigure[Property binning by saccade amplitude.]{\includegraphics[width=0.99\linewidth]{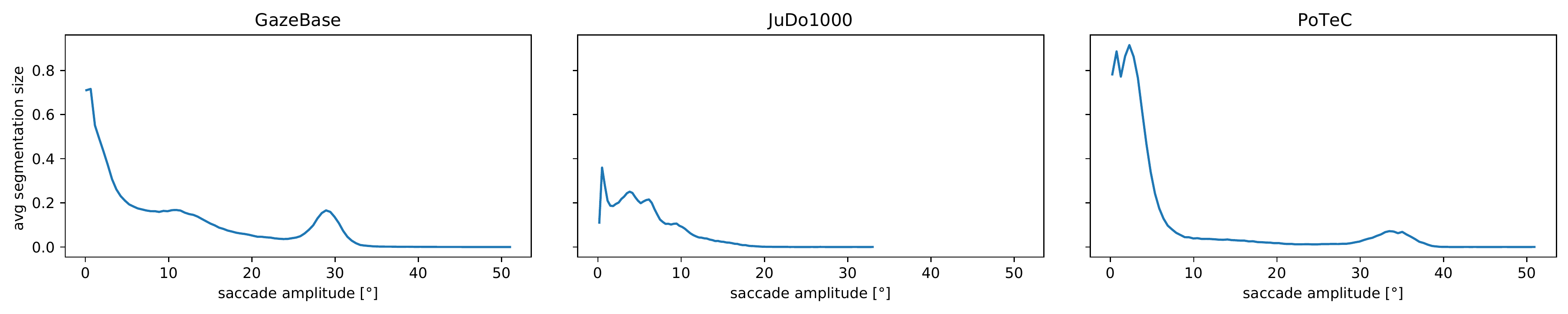}\label{fig:event-property-binning-saccade-amplitude-size}}\\
\subfigure[Property binning by fixation dispersion.]{\includegraphics[width=0.99\linewidth]{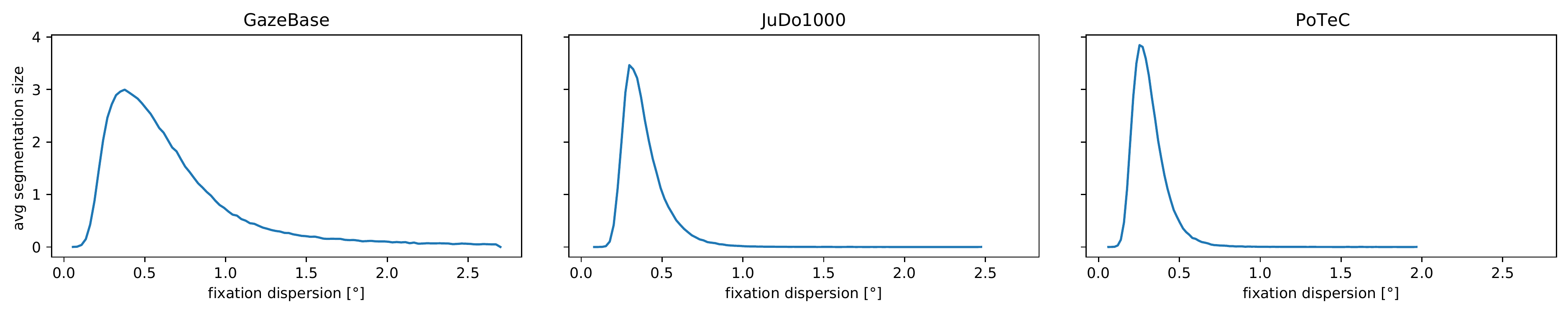}\label{fig:event-property-binning-fixation-dispersion-size}}\\
\subfigure[Property binning by fixation velocity standard deviation.]{\includegraphics[width=0.99\linewidth]{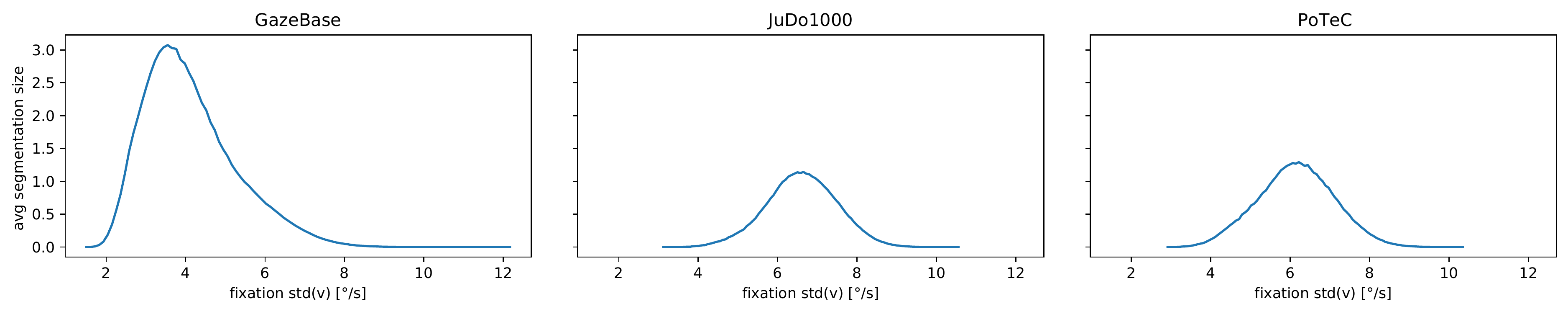}\label{fig:event-property-binning-fixation-vstd-size}}\\
\caption{Segmentation sizes across various event properties.}
\label{fig:event-property-binning-sizes}
\end{figure*}

\begin{figure}
\centering
\includegraphics[width=\linewidth]{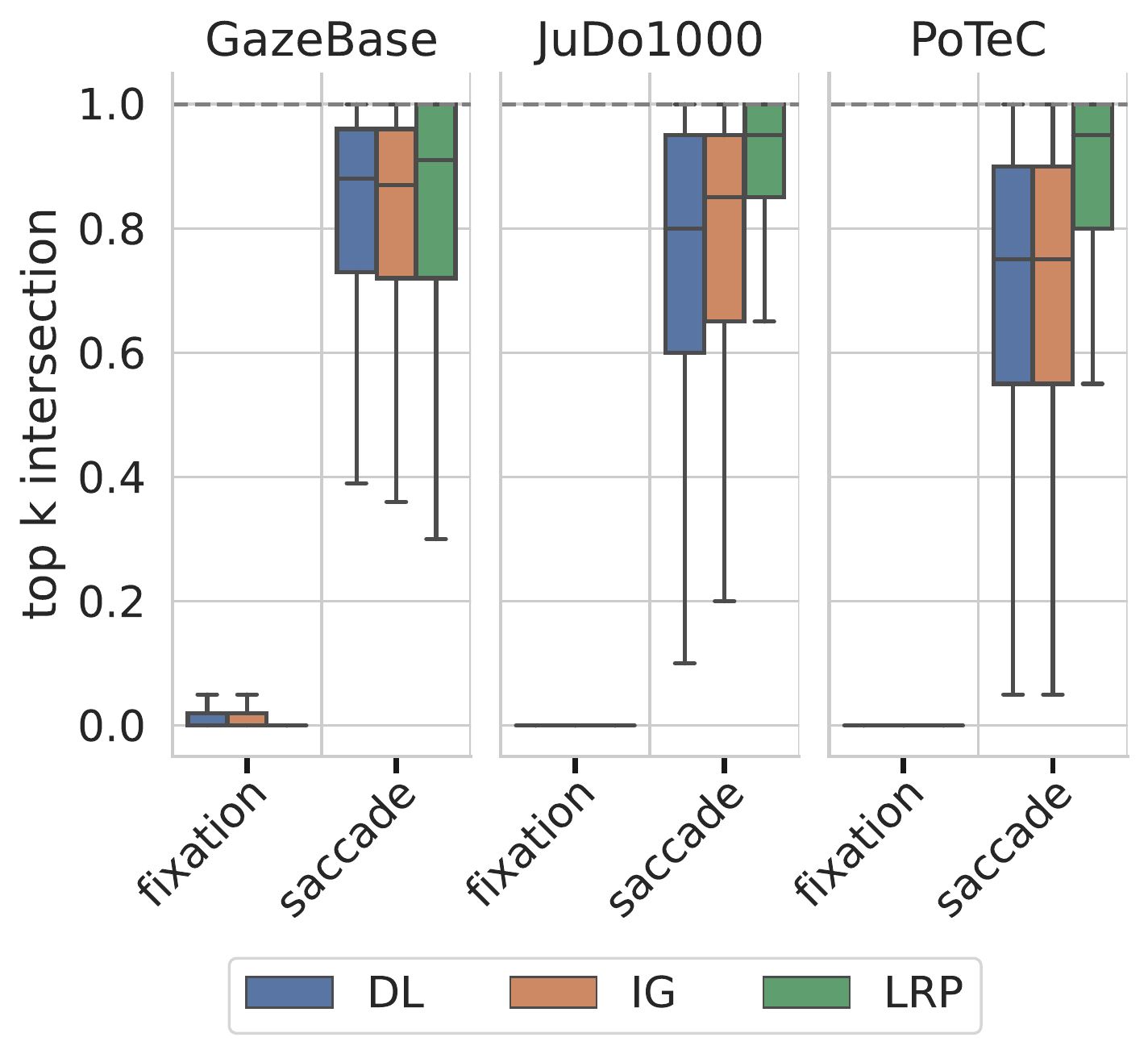}
\caption{Top 2\% intersection of saccades and fixations.}
\label{fig:event-type-tki}
\end{figure}

\begin{figure}
\centering
\includegraphics[width=\linewidth]{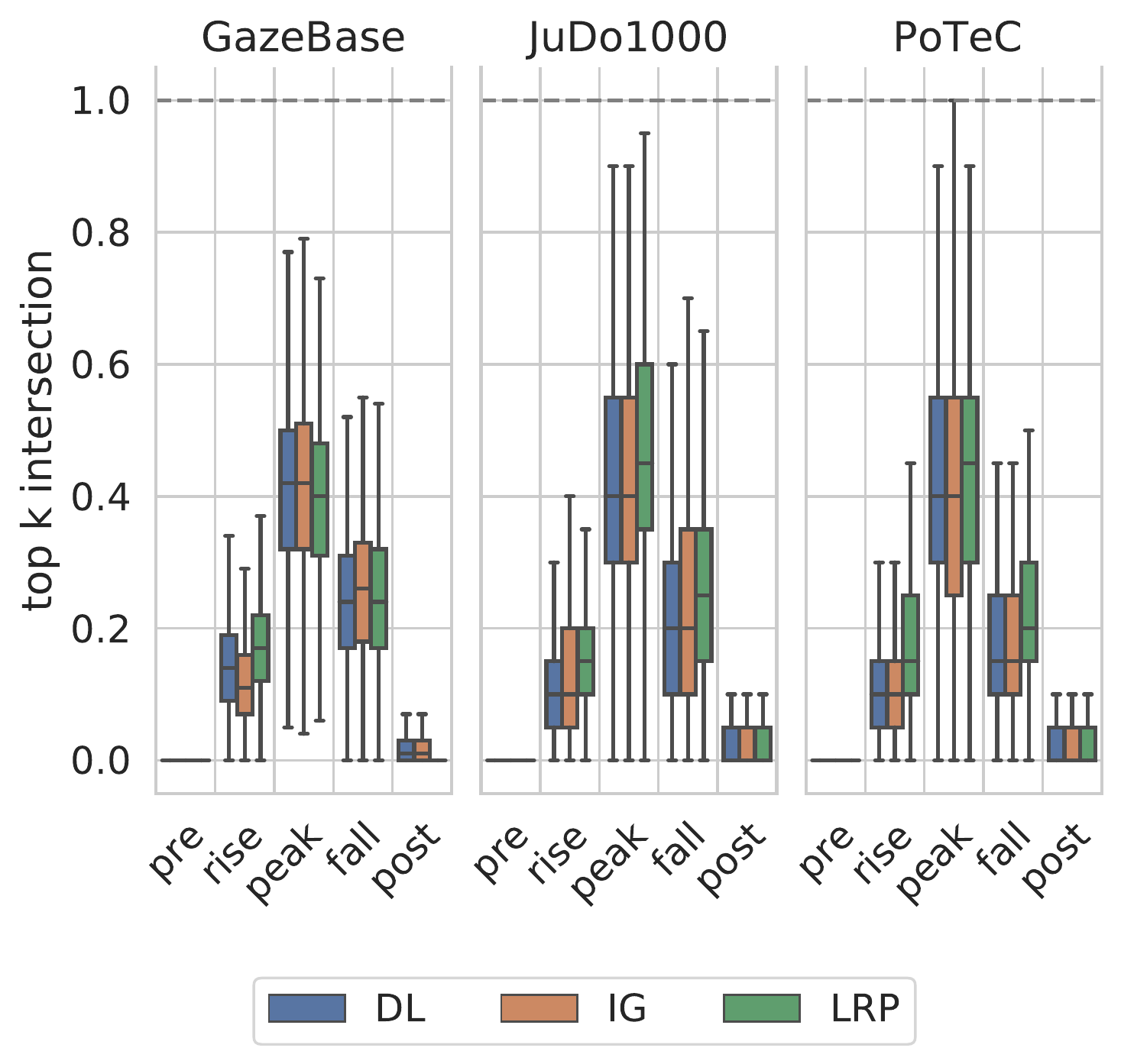}
\caption{Top 2\% intersection for sub-events of saccades.}
\label{fig:event-dissection-tki}
\end{figure}

\section{Top K Intersections}
This section contains the top $k$ intersection values for each of the experiments from the main paper.
$k$ is set to the top 2\% of the attributions.

\subsection{Saccades and Fixations}
\label{apx:event-type-tki}

Figure~\ref{fig:event-type-tki} contains the Top K Intersection values ($K = 2\%$) of saccades and fixations. The values are scaled according to Equation~\ref{eq:ci} to get the results from Figure~\ref{fig:event-type-ci}.

\subsection{Saccade Sub-events}

Figure~\ref{fig:event-dissection-tki} contains the Top K Intersection values ($K = 2\%$) of saccade sub-events. The values are scaled according to Equation~\ref{eq:ci} to get the results from Figure~\ref{fig:event-dissection-ci}.

\clearpage

\end{appendix}

\end{document}